\pdfoutput=1

\documentclass[11pt]{article}

\usepackage[final]{acl}

\usepackage{times}
\usepackage{latexsym}

\usepackage[T1]{fontenc}

\usepackage[utf8]{inputenc}

\usepackage{microtype}

\usepackage{inconsolata}

\usepackage{graphicx}

\usepackage{soul}
\definecolor{lightblue}{rgb}{0.68, 0.85, 0.9}
\sethlcolor{lightblue}

\usepackage{amsmath,amssymb}
\usepackage{dsfont}

\usepackage{geometry}
\usepackage{booktabs}
\usepackage{makecell}
\usepackage{multirow}

\usepackage[most]{tcolorbox}
\usepackage{fontawesome5}
\definecolor{yellow}{HTML}{F6BD60}
\usepackage{multicol}

\usepackage{rotating}
\usepackage{tabularx}

\usepackage{algorithm}
\usepackage{algpseudocode}

\usepackage{colortbl}

\title{Calibrating LLMs with Preference Optimization on Thought Trees for Generating Rationale in Science Question Scoring}

\author{Jiazheng Li$^1$, Hainiu Xu$^1$, Zhaoyue Sun$^{1,4}$\\
{\bf Yuxiang Zhou$^1$, David West$^2$, Cesare Aloisi$^2$, Yulan He$^{1,3}$}\\
  $^1$Department of Informatics, King's College London, UK~~~~~~$^2$AQA, UK\\
  $^3$The Alan Turing Institute, UK~~~$^4$Department of Computer Science, University of Warwick, UK\\
\texttt{\{jiazheng.li, hainiu.xu, yuxiang.zhou, yulan.he\}@kcl.ac.uk}\\
  \texttt{\{caloisi, dwest\}@aqa.org.uk}~~~\texttt{zhaoyue.sun@warwick.ac.uk}}

\begin{document}
\maketitle
\begin{abstract}
Generating rationales that justify scoring decisions has been a promising way to facilitate explainability in automated scoring systems. However, existing methods do not match the accuracy of classifier-based methods. Plus, the generated rationales often contain hallucinated information. To address these issues, we propose a novel framework capable of generating more faithful rationales and, more importantly, matching performance with classifier-based black-box scoring systems. We first mimic the human assessment process by querying Large Language Models (LLMs) to generate a thought tree. We then summarise intermediate assessment decisions from each thought tree path for creating synthetic rationale data and rationale preference data. Finally, we utilise the generated synthetic data to calibrate LLMs through a two-step training process: supervised fine-tuning and preference optimization. Extensive experimental results demonstrate that our framework achieves a 38\% assessment performance improvement in the QWK score compared to prior work while producing higher-quality rationales, as recognised by human evaluators and LLMs. Our work sheds light on the effectiveness of performing preference optimization using synthetic preference data obtained from thought tree paths\footnote{Data and code are available at \url{https://github.com/lijiazheng99/thought_tree_assessment}.}.
\end{abstract}

\section{Introduction}
\begin{figure}[t]
\centering
\includegraphics[width=\linewidth]{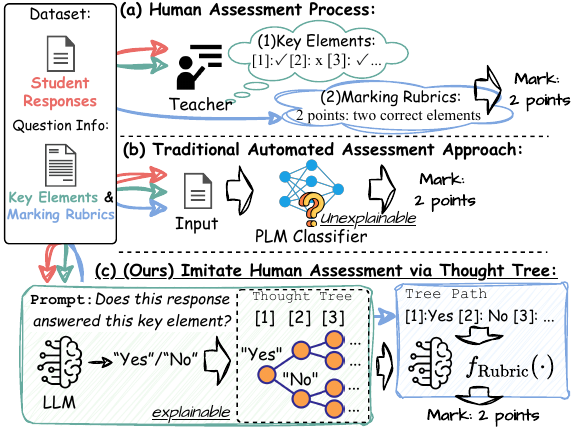}
\caption{Illustration of question information used in (a) Human assessment and (b) Traditional PLM-based text classifier assessment approaches. We propose (c) Imitating the human assessment procedure by prompting LLMs to generate thought trees that form rationales.}
\label{fig:intro}
\end{figure}

Student answer marking requires substantial human efforts. 
The human scoring process (e.g., Figure \ref{fig:intro}(a)) typically involves first understanding the content of the student response, then identifying valid key answer elements (e.g., ``\textit{student should answer what materials to test in their response}''), and finally assigning a score according to a marking rubric (e.g., ``\textit{2 points for accurately describe two additional pieces of information}'') \cite{critical_thinking, assessment_method}. 
Numerous studies \cite{helen-aes-2016, yue-aes-2017} have aimed to automate such an assessment pipeline, thereby reducing labour and standardising assessment criteria.
Traditional automated assessment methods generally address this problem through text classification \cite{grading_classification} or regression \cite{xie-etal-2022-automated}, with the goal of predicting the score for a given student response (Figure \ref{fig:intro}(b)).

Recent studies suggested that Large Language Models (LLMs) could be leveraged to explain student answer scoring. For example, \citet{li-etal-2023-distilling} utilised ChatGPT-generated rationales as noisy supervision signals to train a smaller model for simultaneous student answer scoring and rationale generation.
However, their framework does not match the accuracy of classification baselines and suffers from hallucination problems, leading to unfaithful and inconsistent rationales.
Moreover, it does not support the assessment of key elements in student responses according to the marking rubric.

To address the aforementioned issues, we propose a novel framework to imitate the human scoring process by dividing the scoring of student answers into two steps (Figure \ref{fig:intro}(c)). Initially, an LLM is tasked with comparing each key answer element against a student answer and providing a binary assessment decision (``\emph{Yes}'' or ``\emph{No}'') indicating its presence or absence. These series of intermediate decisions, combined with prompts, ultimately form an assessment thought path \cite{tot,yao2023react}. Subsequently, the assessment outcomes for each key element are aggregated using a marking rubric to derive the final score. This aggregation process can either be executed using explicit arithmetic calculations in Python code or directly by the LLM. Each thought tree paths can be faithfully summarised into synthetic response-level rationales, facilitating explainable automated student answer scoring. %

Furthermore, the thought tree paths could contain trajectory that leads to correct or incorrect assessments. Such positive and negative rationale samples generated by the thought tree can be regarded as preference data, which allows us to conveniently construct a synthetic rationale preference dataset. Therefore, to calibrate LLMs for both accurate assessment of student answers and faithful generation of corresponding rationales, we align LLMs with our preference dataset using advanced RLHF methods \cite{NIPS2017_d5e2c0ad, rlhf}. Specifically, our alignment pipeline contains the following two steps: (1) supervised fine-tuning (SFT) on synthetic \emph{response-level rationale data}, followed by (2) preference optimization \cite{dpo} on synthetic \emph{preference data}. To address data scarcity, we generate synthetic data nearly three times larger than the original dataset. Our experimental results demonstrate that our framework enhances the explainability of the assessment process by generating more faithful rationales while matching or exceeding the accuracy of traditional text classification approaches.

In summary, our contributions are: \textbf{(1)} We proposed a method for generating more faithful assessment rationales by imitating human assessment processes through thought trees; \textbf{(2)} We developed a technique for generating synthetic preference data by modelling preferences based on the correctness of thought tree paths; %
\textbf{(3)} We carried out extensive experiments, demonstrating that our framework achieved a 38\% improvement in QWK compared to the state-of-the-art assessment rationale generation framework while generating more informative and higher-quality rationales.
To the best of our knowledge, this is the first work to perform preference optimization using synthetic preference rationales that are generated via thought tree paths. %

\begin{figure*}[ht]
  \centering
  \includegraphics[width=\linewidth]{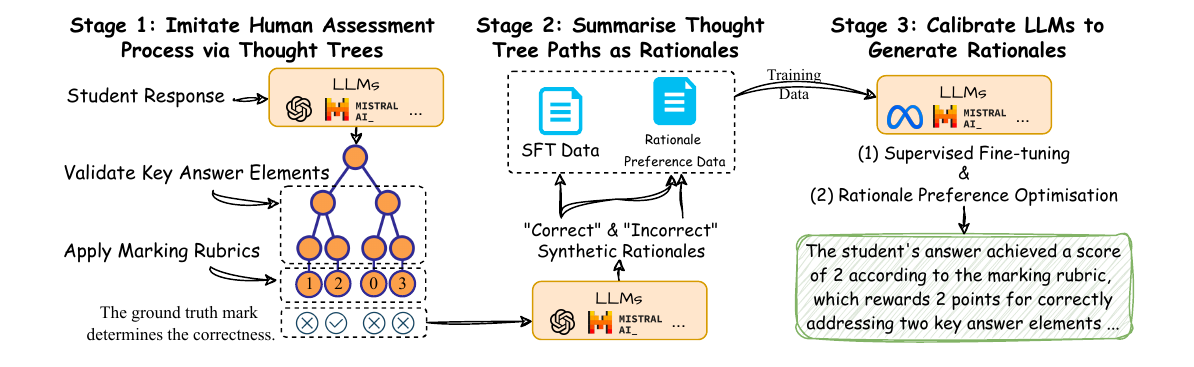}
  \caption{Overview of Our Three-stage Thought Tree Guided Rationale Generation Framework.}
  \label{fig:framework}
\end{figure*}

\section{Thought Tree Guided Rationale Generation Framework}
We propose a \emph{Thought Tree Guided Rationale Generation framework} to improve scoring performance and generate higher-quality assessment rationales. An overview of our proposed three-stage framework is provided in Figure \ref{fig:framework}. In this section, we will first introduce the problem setup and then explain the method used in each stage.

\paragraph{Problem Setup} Datasets for developing automated answer scoring systems typically contain information about questions, such as question prompts, key answer elements, and marking rubrics, along with a set of student responses and the corresponding marks. Formally, let $D = \{(x_i, y_i)\}_{i=1}^N$ represent  $N$ data points under the same question, where $x_i$ denotes a student response and $y_i$ represents the corresponding mark. %

Let $K=\{k_j\}_{j=1}^M$ denote the set of key elements for the current question, where $M$ is the total number of distinct key answer elements. For a points-based marking scheme, the marking rubric can be viewed as a question-dependent, piece-wise scoring function, $f_r(\cdot)$, defined as:
\begin{equation}
y_i = f_r(\mathbf{v}(x_i, K)) \label{formula:real_stuff},
\end{equation}
where $\mathbf{v}(x_i, K) \in \mathbb{R}^{M}$ is a multi-hot vector indicating the coverage of key answer elements $k_j \in K$. 
Specifically, the $j^{th}$ index of $\mathbf{v}$ takes the form of $\mathds{1}_{x_i}(k_{j})$ that returns ``\textit{Yes}'' if the key element $k_{j}$ correctly answered in $x_i$ and ``\textit{No}'' otherwise.

The application of the marking rubric is purely deterministic; therefore, $\mathds{1}_{k_{j}}(\cdot)$ becomes the most critical component influencing the quality of the human assessment process.

\subsection*{Stage 1: Imitate Human Assessment Process via Thought Trees} \label{sec:stage1}
Prior work \cite{li-etal-2023-distilling} on explainable student answer scoring suggests that LLMs may struggle to perform the faithful assessment in one step, which may sometimes lead to hallucinations. To tackle this issue, we propose using LLMs to mimic human assessment decision-making by breaking the overall assessment task into intermediate assessment decisions that form deterministic thought trees.

\begin{figure}[h]
  \centering
  \includegraphics[width=0.90\linewidth]{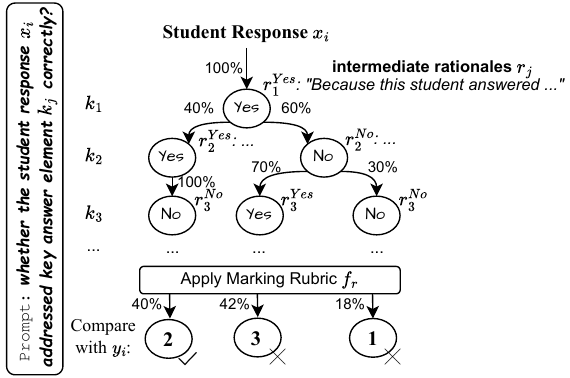}
  \caption{Illustration of the structure of a thought tree.}
  \label{fig:frame_stage1}
\end{figure}

As illustrated in Figure \ref{fig:frame_stage1}, our thought trees contain two levels of intermediate assessment. For the first level, we employ LLMs to imitate the key answer element validation in the human assessment process, denoted as $\mathds{1}_{x_i}(k_{j})$. We constrain the assessment decision for each key element $k_j$ to a binary classification task. Each decision is sampled using Monte Carlo methods to mimic the variation of human decisions \cite{6145622}. Once decisions are obtained for all key answer elements, the final leaf node aggregates the path's final mark $\hat{y}_i$ according to the marking rubric using explicit arithmetic calculations in Python code or directly by LLM.

We formally define our thought tree as follows: given an LLM $\texttt{LLM}_{\theta}$ (e.g., GPT-4), we prompt the LLM with the question information and student response $x_i$ to generate an assessment thought tree. Initially, at the key element assessment level, the assessment decision $z_j$ for a key answer element $k_j$ is obtained by sampling LLM $n$ times:
\begin{equation}
     z_{j}^{(t)} = \texttt{LLM}_{\theta}(x_i, k_j), t=1,2,...,n,\forall k_j \in K,
\end{equation}
\noindent where we use an LLM parameterized by $\theta$ to mimic the human judgement process for identifying key answer elements from the student answer, denoted as $\mathds{1}_{x_i}(k_{j})$.
Prompt messages adhere to a template of ``\textit{whether the student response $x_i$ addressed key answer element $k_j$ correctly}''. Thus, akin to human judgement, the decision $z_j$ can only be either ``\textit{Yes}'' or ``\textit{No}''. We can aggregate similar decisions to derive a sampled decision probability:
\begin{align}
     P(z_{j}^{\text{Yes}}) &= {|\{t : z_{j}^{(t)} = 1\}|/n},\notag\\
     P(z_{j}^{\text{No}}) &= {|\{t : z_{j}^{(t)}  = 0\}|/n},
\end{align}
where $|\cdot|$ denotes the cardinality of the given set. These decision probabilities $P(z_{j}^{\text{Yes}})$ and $P(z_{j}^{\text{No}})$ represent the likelihood of each decision outcome, mimicking variations in real assessment decisions. 

To ensure faithful intermediate decisions for aggregating response-level rationale at a later stage, we generate simpler, key element-wise rationales $r_j$ that justify each unique decision outcome:
\begin{align}
   r_j = \texttt{LLM}_{\theta}(x_i, k_j, z_{j}).
\end{align}
Once every key answer element has been traversed, we can formulate and represent each distinct tree path as a multi-hot encoding using $\mathbf{v}(\cdot)$ from Eq. (\ref{formula:real_stuff}). Assuming we have $d$ distinct paths, where $1\leq d \leq 2^{(M-1)}$, each path can be represented as:
\begin{equation}
   \text{path}_l = \hat{\mathbf{v}}(\mathcal{Z}), l = 1,2,...,d,
\end{equation}
where $\hat{\mathbf{v}}$ is an estimate of $\mathbf{v}$, representing the coverage of key answer elements, and $\mathcal{Z}$ is a set of binary indicators derived from $P(z_j^{\text{Yes}})$ and $P(z_j^{\text{No}})$. The probability of each path \(P(\text{path}_l)\) is the product of decision probabilities along the path:
\begin{equation}
    P(\text{path}_l) = \prod_{j=1}^{M} P(z_j).
\end{equation}
Subsequently, we apply the marking rubric \(f_r\) to obtain a score for each path:
\begin{equation}
    \hat{y}_{\text{path}_{l}} = f_r(\text{path}_l).
\end{equation}
Although the marking rubric is programmatic, it is still provided in text descriptions within the question format. In practical implementation, we utilize the LLM to dynamically transform the textual marking rubric into executable python code, which takes key element-level assessment decisions as a list as input and provides a score as output.

Finally, the predicted score for the current thought tree is determined by the path with the highest probability: 
\begin{align}
\hat{y}_{tree} &= \hat{y}_{\text{path}_{l^*}},\notag\\
&\text{where}\quad l^* = \arg \max_{l} P(\text{path}_l).
\end{align}
For detailed implementation, search algorithm, prompt setup and ablation studies on thought trees, please refer to Appendix \ref{app:stage1_further}.

\subsection*{Stage 2: Summarise Thought Tree Paths as Rationales} \label{sec:stage2}
In the second stage, we employ the thought trees generated in Stage 1 as reliable intermediate assessment decisions for generating synthetic student response-level rationales. To mitigate hallucination and maintain alignment with the thought tree, we leverage the extensive research on generating faithful summaries with LLMs and frame the synthetic rationale generation as a text summarization task \cite{summarization_1, summarization_2, lyu-etal-2024-towards}.

In this stage, we present the LLM with the question, student answer $x_i$, each key answer element $k_j$, key answer element-level assessment decision $z_j$, and rationale $r_j$, along with textual descriptive marking rubric and the predicted path score $\hat{y}_{\text{path}_{l}}$, in a prompt template, and ask the LLM summarize the response-level rationales. To ensure a diverse generation of free-form rationales, we require the LLM to output in JSON format, containing three columns of outputs:
\begin{itemize}
  \setlength{\itemsep}{1pt}
  \setlength{\parskip}{0pt}
  \setlength{\parsep}{0pt}
    \item[] \textbf{Mark}: Output the assessed mark, which should be the same as $\hat{y}_{\text{path}_{l}}$.
    \item[] \textbf{Rationale}: Summarizes a response-level rationale using each key element-level assessment decision and rationales, and explicitly assign a mark applying the marking rubric.
    \item[] \textbf{Suggestion}: If the student didn't receive a full mark, we extract the answer improvement suggestion from the assessment rationale to provide useful improvement feedback.
\end{itemize}
Eventually, we construct the following three datasets using the above-generated resources:
\paragraph{Supervised Fine-tuning (SFT) Data}
In this dataset, we form SFT data by choosing the path with the minimum absolute predicted score difference with the ground truth score: $l^*=\arg\min_l \left| \hat{y}_{\text{path}_{l}} - y_i \right|$. The \textbf{Rationale} and \textbf{Suggestion} columns for summarising the path $l^*$ are used to construct multi-round assessment SFT data. The first round of the assessment is provided with the entire question information, and the student answer $x_i$ to request an assessment rationale. The output is the \textbf{Rationale}. If the student didn't get a full score in the second round, we continue prompting the LLM to provide an improvement \textbf{Suggestion} based on the first round chat history.
\paragraph{Rationale Preference Data}
As we want to generate faithful rationales without compromising assessment performance, we frame assessment correctness as human preferences, where a more accurate assessment rationale is preferred without disrupting the faithfulness and coherence of the rationale \cite{plm_hp, zhang2024chainpreferenceoptimizationimproving,lu2024eliminatingbiasedlengthreliance, xie2024monte}. We then choose the \textbf{Rationale} from correctly predicted $\text{path}_l$, where $\hat{y}_{\text{path}_{l}}= y_i$, as our preferred rationale, denoted as $\hat{y}_{r^+}$. Other incorrectly predicted $\text{path}_l$s, where $\hat{y}_{\text{path}_{l}}\neq y_i$, as not preferred rationales, denoted as $\hat{y}_{r^-}$. 
Therefore, for an input $x$, the preference for the rationale is denoted as $\hat{y}_{r^+} \succ \hat{y}_{r^-} | x$. The preferences are assumed to be generated by some latent reward model $r^{*}(y, x)$. Subsequently, we use the Bradley-Terry \cite{bt} model to hypothesizes the rationale preference distribution $p^{*}$ can be estimated via pairwise comparison:
\begin{align}
    p^{*}(\hat{y}_{r^+} &\succ \hat{y}_{r^-} \mid x) = \notag \\
    &\frac{\exp(r^{*}(x, \hat{y}_{r^+}))}{\exp(r^{*}(x, \hat{y}_{r^+})) + \exp(r^{*}(x, \hat{y}_{r^-}))}. 
    \label{eq:preference_modelling}
\end{align}

\paragraph{Rationale-to-Score Data}
Different from \citet{li-etal-2023-distilling}, rationales generated from our framework are in a free-text format without any template constraints. Thus, evaluating the assessment performance of rationales becomes a challenge. To automatically extract the mark from the rationale, we construct a dataset using \textbf{Rationale} as input and \textbf{Mark} as output, which is used to train a score extractor for assessing performance.

Questions sometimes involve parallel key elements, where different paths may result in the same score, and all could be considered correct. To ensure clarity and relevance, we present the question details, student responses, and all generated rationales to the LLM. Then, it is directed to resolve the conflict by selecting the most appropriate and accurate rationale that reflects the student's response, after which all other rationales are discarded.

We provide prompt templates, generated data statistics, and used instructions in Appendix \ref{app:stage2_further}.

\subsection*{Stage 3: Calibrate LLMs to Generate Rationales}
Finally, we utilize the synthetic rationale data generated from Stage 2 to calibrate LLMs for rationale generation via two steps of training: 

\paragraph{Supervised Fine-tuning} We fine-tune an LLM on the synthetic SFT dataset with the standard cross-entropy loss to obtain a model $\pi_{\text{SFT}}$. %
\paragraph{Preference Optimization on Rationales}
We parameterize the preference in Eq. (\ref{eq:preference_modelling}) into a reward model $r_{\phi}(x, y)$ and estimate the parameters via maximum likelihood. This problem can be framed as a binary classification task with a negative log-likelihood loss:
\begin{align}
    \mathcal{L}_R(r_\phi, D) &= -\mathbb{E}_{(x,\hat{y}_{r^+},\hat{y}_{r^-}) \sim D} \left[\log\left(\sigma(r_\phi(x, \hat{y}_{r^+})\right.\right. \notag\\
    &\left.\left.- r_\phi(x, \hat{y}_{r^-}))\right)\right].
\end{align}
Following Direct Preference Optimization (DPO) \cite{dpo}, we can formulate a maximum likelihood objective for a parameterized policy $\pi_{\theta}$. The policy objective becomes:

\begin{small}
\begin{multline}
    \mathcal{L}_{DPO}(\pi_\theta; \pi_{\text{ref}}) = \\
    -\mathbb{E}_{(x,\hat{y}_{r^+},\hat{y}_{r^-}) \sim D} \left[ \log \sigma \left( \beta \log \frac{\pi_\theta(\hat{y}_{r^+} \mid x)}{\pi_{\text{ref}}(\hat{y}_{r^+} \mid x)} \right.\right. \notag\\
    \left.\left. - \beta \log \frac{\pi_\theta(\hat{y}_{r^-} \mid x)}{\pi_{\text{ref}}(\hat{y}_{r^-} \mid x)} \right) \right],
\end{multline}
\end{small}

\noindent where the $\pi_{\text{ref}}$ is the $\pi_{\text{SFT}}$. Empirically, training solely with $\mathcal{L}_{DPO}$ may cause model degeneration. In our implementation, we add a cross-entropy loss on the preferred rationale $\hat{y}_{r^+}$ to mitigate this issue.

\section{Experiments}
\subsection{Experimental Setup}
\paragraph{Datasets} We use two data sources for our experiments: (1) The Hewlett Foundation Short Answer Scoring (ASAP) dataset, which focuses on short essay questions spanning topics in science and biology. (2) A private dataset comprising student responses with human-assigned scores for biology exam questions from a reputable examination service. We present our dataset statistics in Table \ref{tab:data_statistic}.

\begin{table}[!h]
\centering
\resizebox{\linewidth}{!}{
\begin{tabular}{lcccc}
\toprule
\textbf{Datasets} & Train & Validation & Test & Score Range\\
\midrule
ASAP 1    & 1,337 & 331  & 554 & 0-3 \\
ASAP 2    & 1,018 & 252  & 426 & 0-3 \\
ASAP 5    & 1,436 & 359  & 598 & 0-3 \\
ASAP 6    & 1,437 & 359  & 599 & 0-3 \\ 
Private 1 & 440   & 89   & 254 & 0-4 \\
Private 2 & 358   & 72   & 196 & 0-3 \\
\bottomrule
\end{tabular}}
\caption{Dataset Statistics}
\label{tab:data_statistic}
\end{table}

\paragraph{Baselines}
We compare our method with two baseline setups: \underline{Text Classifier}: an assessment classifier built on DeBERTa Large \cite{debertav3}, designed to assess student answers and output scores \cite{bert_classifer_aes}. \underline{AERA}, an explainable student answer scoring approach \cite{li-etal-2023-distilling}, which distils ChatGPT refined rationales into a LongT5 \cite{longt5} model.

\paragraph{Evaluation Metrics}
We evaluate the assessment performance using Accuracy (Acc), macro F1 score (F1), and Quadratic Weighted Kappa (QWK). 

We provide detailed hyper-parameters setup, model version details, definition of Quadratic Weighted Kappa, etc., in Appendix \ref{app:further_setup}.

\begin{table*}[ht]
\centering
\resizebox{\textwidth}{!}{%
\begin{tabular}{lcccccccccccc}
\toprule
& \multicolumn{3}{c}{\textbf{Private 1} (Biology)} & \multicolumn{3}{c}{\textbf{Private 2} (Biology)} & \multicolumn{3}{c}{\textbf{ASAP 1} (Science)} & \multicolumn{3}{c}{\textbf{ASAP 2} (Science)} \\ \cmidrule(r){2-4} \cmidrule(l){5-7} \cmidrule(l){8-10} \cmidrule(l){11-13} 
\textbf{Methods} & Acc & F1 & \textbf{QWK} & Acc & F1 & \textbf{QWK} & Acc & F1 & \textbf{QWK} & Acc & F1 & \textbf{QWK} \\
\midrule
Text Classifier (C) & 0.6787 & 0.6784 & \underline{0.8853} & 0.6224 & 0.6355 & \underline{0.8385} & 0.7767 & 0.7805 & 0.8528 & 0.6798 & 0.6817 & \underline{0.8187} \\
AERA (R) & 0.4055 & 0.3838 & 0.7139 & 0.2551 & 0.1020 & 0.0024 & 0.6326 & 0.6290 & 0.7506 & 0.4327 & 0.4235 & 0.5415 \\\midrule
\multicolumn{13}{c}{\emph{Our Framework}} \\\midrule
\textit{Thought Tree} (C)& 0.5276 & 0.5078 & 0.7891 & 0.5459 & 0.5362 & 0.6710 & 0.6011 & 0.6145 & 0.7419 & 0.6746 & 0.6852 & 0.7856 \\
\textit{Mixtral 8$\times$7B SFT} (R)& 0.5906 & 0.5799 & 0.8326 & 0.5255 & 0.5213 & 0.7324 & 0.7310 & 0.7398 & 0.8511 & 0.5915 & 0.5837 & 0.6886 \\
\textbf{\textit{Mixtral 8$\times$7B DPO}} (R)& 0.6063 & 0.6016 & \textbf{0.8440} & 0.5867 & 0.5917 & \textbf{0.7802} & 0.7744 & 0.7808 & \textbf{\underline{0.8749}} & 0.6526 & 0.6763 & \textbf{0.7788} \\
\midrule
& \multicolumn{3}{c}{\textbf{ASAP 5} (Biology)} & \multicolumn{3}{c}{\textbf{ASAP 6} (Biology)} & \multicolumn{3}{c}{\textbf{Overall}} & \multicolumn{3}{c}{\textbf{QWK Statistics}} \\ \cmidrule(r){2-4} \cmidrule(l){5-7} \cmidrule(l){8-10} \cmidrule(l){11-13} 
\textbf{Methods} & Acc & F1 & \textbf{QWK} & Acc & F1 & \textbf{QWK} & Acc & F1 & \textbf{QWK} & $\sigma$ & \multicolumn{2}{c}{Cf. w/ AERA}\\
\cmidrule(l){1-13}
Text Classifier (C)& 0.8625 & 0.6055 & 0.8187 & 0.8891 & 0.6118 & \underline{0.8426} & 0.7515 & 0.6656 & \underline{0.8428} & 0.0006 & \multicolumn{2}{c}{+41.21\%} \\
AERA (R) & 0.8378 & 0.5329 & 0.7644 & 0.8937 & 0.6038 & 0.8081 & 0.5762 & 0.4458 & 0.5968 & 0.0933 & \multicolumn{2}{c}{ - } \\\cmidrule(l){1-13}
\multicolumn{13}{c}{\emph{Our Framework}} \\ \cmidrule(l){1-13}
\textit{Thought Tree} (C)& 0.8043 & 0.5317 & 0.7834 & 0.7496 & 0.4429 & 0.5831 & 0.6505 & 0.5530 & 0.7257 & 0.0069 & \multicolumn{2}{c}{ +21.59\%  } \\
\textit{Mixtral 8$\times$7B SFT} (R)& 0.8579 & 0.5948 & 0.8174 & 0.8564 & 0.5942 & 0.7735 & 0.6922 & 0.6023 & 0.7826 & 0.0040 & \multicolumn{2}{c}{ +31.13\% } \\
\textbf{\textit{Mixtral 8$\times$7B DPO}} (R)& 0.8779 & 0.6326 & \textbf{\underline{0.8447}} & 0.8715 & 0.6011 & \textbf{0.8245} & 0.7282 & 0.6473 & \textbf{0.8245} & 0.0015 & \multicolumn{2}{c}{+38.15\%} \\ 
\cmidrule(l){1-13}
\end{tabular}
}
\caption{Comparison of performance on (C)lassification and (R)ationale assessment approaches. The highest QWK has been highlighted in \textbf{bold} for the highest approaches to rationale generation , and an \underline{underline} denotes the highest across all approaches. \textbf{QWK Statistics} provide insight into overall assessment performance: $\sigma$ indicates the QWK score variation across all datasets, with lower values showing better generalization. Cf. w/ AERA compares overall performance with AERA, where higher values denote significant performance improvements.}
\label{tab:model_comparison}
\end{table*}

\subsection{Overall Performance} \label{sec:overall_assessment}
To examine the scoring performance of our proposed framework, Table \ref{tab:model_comparison} summarizes the scoring results of classification-based baselines, the state-of-the-art explainable scoring model, compared with results produced from our framework\footnote{All the scores from free-form rationales are extracted with a score extractor described in Appendix \ref{app:score_extractor}.}.

\noindent\underline{Text Classifier Vs. AERA}: We replicated the AERA framework on our private datasets with ChatGPT\footnote{We didn't upgrade the backend for rationale generation to GPT-4 due to the incompatibility with the refinement module. An analysis of this issue is presented in Appendix \ref{app:aera_model_choice}.}. The results show that AERA is significantly affected by data scarcity (on Private 2). With only a few hundred data points, the model struggles to generalize. Moreover, a comparison on the ASAP datasets reveals a significant performance gap, particularly on ASAP 2. Although AERA offers better explainability, its performance shows a considerable trade-off, with an average 23\% drop in accuracy and a 33\% reduction in F1 score compared to the text classifier.

\noindent\underline{Thought Tree Vs. Baselines}: Thought Trees\footnote{Thought Tree employs GPT-4 as the backend. Results using open-source models can be found in Appendix \ref{app:stage1_further}} are used for generating response-level synthetic rationales. Simply deriving the scoring decisions from the Thought Tree's highest probability paths reduces the performance gap with text classifiers to 14\%. However, as discussed in \textsection{\ref{sec:stage1}}, there is a chance that paths with lower probabilities could also be correct, suggesting there is room for model calibration. Notably, Thought Trees already achieve better assessment results than AERA on Private 1 \& 2 and ASAP 2 \& 5 datasets. %

\noindent\underline{Mixtral 8$\times$7B SFT Vs. Baselines}: %
By fine-tuning Mixtral 8$\times$7B using our synthesized rationales summarized from Thought Trees, we observe a notable improvement over AERA by nearly 20\% in QWK averaged over all datasets, closing the performance gap with text classifiers. 

\begin{figure}[h]
  \centering
  \includegraphics[width=\linewidth]{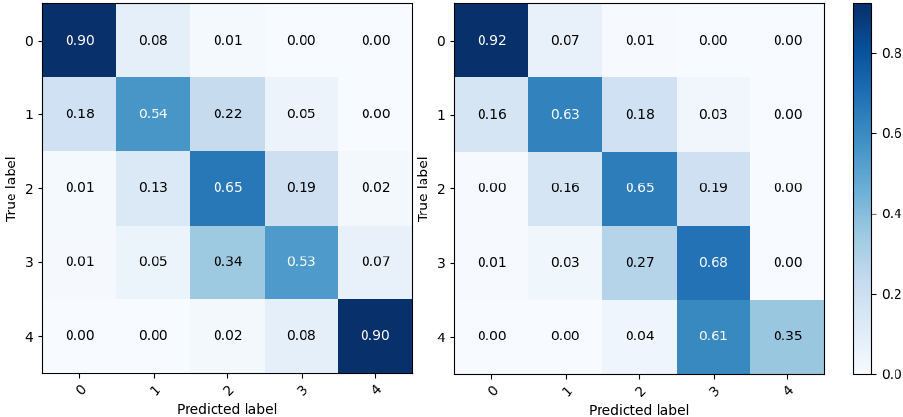}
  \caption{Normalized confusion matrix for Mixtral 8$\times$7B SFT (left) and Mixtral 8$\times$7B DPO (right).}
  \label{fig:sft_dpo_confusion_matrix}
\end{figure}

\noindent\underline{Mixtral 8$\times$7B DPO Vs. All}: The DPO training step shows improved performance across all metrics compared to the SFT stage (nearly 5\% improvement in QWK) and baselines. As shown in Figure \ref{fig:sft_dpo_confusion_matrix}, we compare the confusion matrix between the assessment results from the SFT and DPO steps. Following the DPO training step, all scores ranging from 0 to 3 were better calibrated than the SFT result, showing modelling accuracy as synthetic human preferences could effectively enhance scoring performance. The decline in the score of 4 is attributed to a shortage of data, as Private 1 is the only dataset with a maximum score of 4 with only 51 instances. Remarkably, our DPO step's scoring performance not only outperforms AERA but also matches or exceeds the Text Classifier on the ASAP 1 \& 5 datasets, confirming the effectiveness of our framework in enhancing explainability without compromising scoring accuracy. 

\subsection{Rationale Quality}
Since the dataset lacks ground-truth annotations for rationales, we perform an automated evaluation using LLMs and a human evaluation to examine the quality of generated rationales.

\subsubsection{Automated Evaluation with GPT-4}
To compare the informativeness and correctness of rationales generated by our framework with the AERA baseline, we conduct an automated evaluation using GPT-4. %
The preferred rationale for each pairwise comparison was determined based on which rationale more accurately assessed the student response.  We randomly sampled 30 generated rationales from each dataset, using the same SFT and DPO results presented in Table \ref{tab:model_comparison}. As demonstrated in Table \ref{tab:win_rate}, GPT-4 prefer rationales generated from the SFT model over AERA because they were more informative and directly quoted the student responses. The SFT rationales that were not preferred often incorrectly assessed student responses. A similar preference trend was observed in the comparison between the rationales generated by AERA and DPO. However, the $100\%$ preference rate in the comparison between SFT and DPO does not necessarily imply that DPO is superior. GPT-4 noted that in some cases, both rationales were identical in content, structure, and explanation, stating, ``\textit{Neither is more accurate or helpful than the other, as they provide the same level of detail and critique.}''
\begin{table}[h]
\centering
\resizebox{0.75\linewidth}{!}{%
\small{
\begin{tabular}{lc}
\toprule
\textbf{Pairwise Comparison} & \textbf{Win Rate}\\ 
\midrule
SFT vs. AERA & $99.17\%$ \\
DPO vs. AERA & $100\%$ \\
DPO vs. SFT & $100\%$ \\
\bottomrule
\end{tabular}}}
\caption{Rationale Preferences Evaluated by GPT-4.}
\label{tab:win_rate}
\end{table}

\subsubsection{Human Evaluation}
To evaluate the correctness of assessment decisions based on rationales in more detail, we conducted a human evaluation focusing on the \emph{correctness of matching key answer elements} and the \emph{faithfulness of applying marking rubrics}. We randomly sampled 30 cases from all datasets and provided evaluation instructions to our annotators\footnote{Detail setup is presented in Appendix \ref{sec:human_eval}.}.

As shown in Table \ref{tab:human_evaluation}, our evaluation reveals that rationales generated by our framework more accurately match key answer elements and apply the marking rubric more faithfully. This verifies the effectiveness of the thought tree rationale generation approach, which produces higher-quality rationales. In contrast, some AERA cases showed that the model failed to generate meaningful content, instead repeating template phrases, suggesting difficulties in generalization with their approach.
\begin{table}[h]
\centering
\resizebox{\linewidth}{!}{%
\small{
\begin{tabular}{lccc}
\toprule
\textbf{Evaluated Aspects} & \textbf{DPO} & \textbf{AERA} \\ 
\midrule
Key Element Correctness & \textbf{83.33\%} & 36.67\%\\
Marking Rubric Faithfulness & \textbf{96.67\%} &  66.67\% \\
Accuracy on Sampled Set & \textbf{83.33\%} & 50.00\%\\
\bottomrule
\end{tabular}}}
\caption{Human Evaluation on Rationale Quality.}
\label{tab:human_evaluation}
\end{table}

Interestingly, we found no false positive cases in our framework, meaning the rationales were always meaningful when giving a correct score. However, three such cases were identified in the rationales generated by  AERA. In the sampled DPO cases, incorrect assessments were all attributable to erroneous matching of key answer elements, often caused by overanalysis by the LLM based on context. For example, the LLM incorrectly inferred that the response correctly addressed “\emph{global warming}” based on the mention of “\emph{higher temperature of water}”. This highlights the need for more explicitly defined scopes for the key answer elements. Among the sampled DPO cases, only one was found to apply the marking rubric incorrectly. Therefore, the majority of errors stem from mismatching key elements.

\begin{table*}[!h]
\centering
\resizebox{\linewidth}{!}{%
\begin{tabular}{p{1.02\linewidth}}
\hline
\small
\textbf{Student Response} (from ASAP 2): The pour into the take the vinegar in each containers.  who is the problem (Hypothesis)  How you can do this experiment A table who explain the experiment use more types of masses like glass, paper, etc. \\
\small \textbf{Ground-truth mark}: 0\\
\small \textbf{AERA (A)}: 1 point; This response describes one additional piece of information that would be needed to accurately replicate the experiment:  \textcolor{red}{`the pour into the water in each container.'}\\
\small \textbf{Mixtral 8$\times$7B DPO (B)}: The student's answer scored 0 points as per the marking rubric, which requires the response to describe additional pieces of information needed to accurately replicate the experiment. The student's answer failed to specify any necessary details such as the amount of vinegar used, the type of vinegar, the materials tested, the size/surface area of materials, the rinsing duration, the drying time, the drying method, or the size/type of container.  \textcolor{red}{The student's response included vague and unrelated statements such as `The pour into the take the vinegar in each containers. who is the problem (Hypothesis) How you can do this experiment A table who explain the experiment use more types of masses like glass, paper, etc.'} which do not address the specifics required for replication.\\
\small \textbf{GPT-4 Evaluation}: Comparison: Rationale B is more accurate and helpful because  \textcolor{red}{it clearly explains why the student's response failed to meet the requirements of the marking rubric by not providing specific and relevant information needed to replicate the experiment}, whereas Rationale A incorrectly credits the student for providing useful information. More accurate and helpful: B\\
\small \textbf{Mixtral 8$\times$7B SFT}: The student's answer scored 0 points as per the marking rubric, which requires the response to describe additional pieces of information needed to accurately replicate the experiment. The student's answer failed to specify any necessary details such as the amount of vinegar used, the type of vinegar, the materials tested, the size/surface area of materials, the rinsing duration, the drying time, the drying method, or the size/type of container. The student's response included vague and unrelated statements such as `The pour into the take the vinegar in each containers. who is the problem (Hypothesis) How you can do this experiment A table who explain the experiment use more types of masses like glass, paper, etc.' which do not address the specifics required for replication.  \emph{(Exactly the same as DPO output)}\\
\hline
\small \textbf{Student Response} (from ASAP 5): mRNA will change into tRNA and will leave amino acids. Then the amino acids will make a ribosome. mRNA will send a message to tRNA. Finally ribosomes will connect and make a bond \quad\textbf{Ground-truth Mark}: 0\\
\small \textbf{AERA (A)}: 1 point; The student answer matches one key element, \textcolor{red}{`...mRNA is sent to the ribosomes...'}. However, the other statements are incorrect or incomplete.\\
\small \textbf{Mixtral 8$\times$7B SFT (B)}: The student's answer scored 0 points according to the marking rubric which awards 3 points for addressing four key elements, 2 points for three key elements, 1 point for one or two key elements, and 0 points for other responses. \textcolor{red}{The student's answer contains several fundamental errors and misunderstandings about the protein synthesis process. Key inaccuracies include stating`mRNA will change into tRNA' and `the amino acids will make a ribosome,' which demonstrate a confusion between the roles and identities of mRNA, tRNA, and ribosomes.} The student also fails to describe critical steps such as mRNA exiting the nucleus, mRNA traveling to the ribosome, the reading of mRNA codons by rRNA, and the specific interactions and roles of tRNA in carrying and adding amino acids. None of the key answer elements were correctly addressed as required by the marking rubric.\\
\small \textbf{GPT-4 Evaluation}: Comparison: Rationale B is more detailed and accurate, \textcolor{red}{providing specific feedback on the student's misunderstandings and errors, which is essential for learning}, whereas Rationale A is overly generous and lacks detail on the specific errors made. More accurate and helpful: B\\
\small \textbf{Mixtral 8$\times$7B DPO}: The student's answer scored 0 points according to the marking rubric, which awards 0 points if none of the key elements are correctly addressed. \textcolor{red}{The student's statements such as`mRNA will change into tRNA' and `the amino acids will make a ribosome' demonstrate a fundamental misunderstanding of mRNA, tRNA, and ribosome roles and functions in protein synthesis.} None of the required key answer elements, including mRNA exiting the nucleus, mRNA traveling to the ribosome, reading of codons, and interactions between tRNA and mRNA, were correctly described.\\
\hline
\end{tabular}}
\caption{Case studies on rationale quality. Critical statements from the rationales are highlighted in \textcolor{red}{red}.}
\label{tab:case_study}
\end{table*}

\subsection{Case Studies for Rationale Quality}

As demonstrated in Table \ref{tab:case_study}, we provide several case studies to compare rationales generated by the AERA, Mixtral 8$\times$7B SFT, and DPO models.

The first example is from the ASAP 2 dataset, where the correct mark is 0. The AERA model incorrectly assessed the student's answer, quoting \textit{``the pour into the water in each container''}, which was not a valid key answer element. In contrast, the DPO output correctly assigned a score of 0 and highlighted the vagueness of the student's statements, directly quoting the exact student's answer. This direct quoting enhances the faithfulness of the generated rationale. The usefulness of the DPO's rationale is further recognized by GPT-4, noting that rationale B is more informative. Interestingly, the DPO-generated rationale remains consistent with the SFT stage output; possibly, it already aligns with the model's preferred distribution.

The second case involves a student response from ASAP 5, where the correct mark is 0. Again, the AERA model incorrectly inferred the key answer element ``mRNA is sent to the ribosomes'' from the student's answer. In contrast, the SFT model correctly identified the student's fundamental errors and misunderstandings about the process, providing useful and specific feedback, which GPT-4 also recognized. The Mixtral 8$\times$7B DPO output provides almost the same information while being more concise, indicating that the DPO stage effectively removes redundant information while reinforcing assessment accuracy.

Further extensive experiments, case studies and error analyses are provided in Appendix \ref{app:further_experiments}.
\section{Related Work}
\paragraph{Automated Student Answer Scoring} Existing automated student answer scoring (ASAS) systems typically frame the problem as a text classification task \cite{grading_classification, taghipour-ng-2016-neural}. Most ASAS research focuses on essay scoring, evaluating essays based on language quality \cite{esol_dataset, grammar_correction, velentzas-etal-2024-logging-keystrokes}. Various approaches have aimed to enhance the trustworthiness of these systems by explaining the decision-making processes through feature analysis \cite{dong-zhang-2016-automatic, yang-etal-2021-exploring, bert_feature, li-uncertainty-interpretation} or visualizing weights and self-attention mechanisms \cite{helen-aes-2016,yang-etal-2020-enhancing}. However, due to limited resources, the scoring and explanation of science questions have not been extensively studied. 
\citet{li-etal-2023-distilling} proposed a rationale generation framework for science question scoring, but their approach suffers from hallucination issues and an apparent trade-off in assessment performance. Unlike previous works, our framework managed to generate more accurate assessment rationales that are more faithful to the assessment material.

\paragraph{Intermediate Rationale for LLM Problem Solving} Recent developments in LLMs have demonstrated advances in reasoning capabilities \cite{zero_shot, scaling_law, few_shot, zhou2024mysteryincontextlearningcomprehensive, zhu-etal-2024-explanation}. However, complex reasoning in a single step remains challenging for LLMs \cite{zhang2024prompting}. Researchers proposed CoT \cite{cot}, which aimed at breaking down complex reasoning into intermediate steps. Building on this idea, various prompting methods have been proposed to improve the success rate in solving complex tasks, such as Self-Consistency \cite{selfconsistency}, which utilizes the notion of bagging, and Tree-of-Thought (ToT) \cite{tot,zhang2024prompting}, which increases the breadth of search on top of CoT. In our framework, we propose a thought tree to imitate the human assessment process via breaking down assessment task into intermediate assessment decisions. Our thought tree differs from the ToT in two significant ways: (1) Our task requires precise planning. Instead of prompting LLMs to generate plans, prompts in our intermediate steps are determined by each key element. Consequently, our tree has a fixed maximum depth of $M+1$ and width of $2^{(M-1)}$ for $M$ key answer elements. (2) The goal of the ToT is to find the most promising states that solve the problem more efficiently, whereas our goal is to sample all available paths, regardless of correctness, and generate more data for training.

\section{Conclusion}
In this work, we introduced a novel Thought Tree Guided Rationale Generation Framework on providing more accurate scoring decisions justified by more faithful rationales. 
We employed a thought tree to imitate human assessment decision-making to enable more faithful intermediate assessment. Each tree path was then summarized into synthetic rationales. 
Extensive experiments, LLM evaluation, and human evaluation results showed that calibrating LLMs with our framework-generated synthetic response-level rationales and synthetic rationale preference data can effectively provide more informative rationales that align with student responses while achieving higher scoring performance.
Our work highlights the potential for preference optimization using synthetic preference data obtained from thought tree paths.

\section*{Limitations}
This study has several limitations: \textbf{Complexity of Tree Paths}: Tree query presented in the first stage of our framework may take huge computational resources if the set of key answer elements is large and the backend LLM is incapable of doing assessment (e.g. the thought tree could have $2^{(M-1)}$ distinct paths in the worst case). \textbf{Generation of Data Limited by LLM Content Filter}: Many language models have been aligned to prevent harmful use. Filtering harmful content may influence the data production rate in our framework. We noticed a few cases in which student responses were not harmful, and answers about the ``death of plants'' or ``extinction of animals'' eventually were refused assessment by LLMs. This limitation is more dataset-specific and only happens to questions that contain ambiguous terms and need to rely on the context. \textbf{Noisy Synthetic Rationales}: Although we have seen promising improvement in the assessment performance and rationale quality, the synthetic rationales generated from the second stage of our framework are still noisy. However, tackling this issue would require extensive human verification. %

\section*{Ethics Statement}
The datasets utilized in this study contain an open-source and a non-open-source collection of anonymous student responses and do not contain any sensitive or identifiable information. We carried out a screening on LLM outputs and didn't observe any outputs containing harmful information or exposing personal information. However, our framework needs careful evaluation by experts on the assessment decisions and rationales before being put into a high-stakes examination assessment environment.

\section*{Acknowledgements}

We thank anonymous reviewers for their reviews. This work was supported in part by the UK Engineering and Physical Sciences Research Council through a Turing AI Fellowship (grant no. EP/V020579/1, EP/V020579/2) and Innovate UK through the Accelerating Trustworthy AI programme (grant no. 10093055). JL is funded by a PhD scholarship provided by AQA, and OpenAI Researcher Access Program.

\bibliography{custom}

\clearpage

\appendix
\setcounter{table}{0}
\renewcommand{\thetable}{A\arabic{table}}
\setcounter{figure}{0}
\renewcommand{\thefigure}{A\arabic{figure}}

\section{Further Experimental Setup}
\label{app:further_setup}
\subsection{Further Dataset Statistics}
We utilize two datasets, The Hewlett Foundation: Short Answer Scoring (ASAP-SAS) dataset \cite{asap-aes} and our private dataset. Detailed statistics of the dataset are presented in Table \ref{tab:data_statistic}. Following the setup from \cite{li-etal-2023-distilling}, for the ASAP dataset, we focus on non-objective science and biology questions. Please note that due to GPT-4's filtering of harmful content, one instance from the original ASAP 5 dataset has been removed from the training set.
\subsection{Hyper-parameters and Model Versions}
We report all the parameters and configurations we used in Table \ref{tab:hyper-parameters}.
\begin{table*}[htbp]
\centering
\resizebox{0.8\linewidth}{!}{%
\begin{tabular}{llll}
\toprule
\multicolumn{4}{c}{\textbf{Stage 1}}\\
\midrule
\textbf{Model Versions}: &  \\
\multicolumn{2}{l}{Gpt-4 API as tree backend} & \multicolumn{2}{l}{\texttt{gpt-4-1106-preview}} \\
\multicolumn{2}{l}{Mixtral 8$\times$7B API as tree backend} & \multicolumn{2}{l}{\texttt{open-mixtral-8x7b}} \\
\multicolumn{2}{l}{Mixtral 8$\times$7B as tree backend} & \multicolumn{2}{l}{\texttt{mistralai/Mixtral-8x7B-v0.1}} \\
\multicolumn{2}{l}{\textbf{Generation Configs}:} &  \\
\texttt{temperature} & \texttt{0.7} & & \\
\texttt{candidates} for decision & \texttt{10} & \texttt{candidates} for rationale & \texttt{1} \\
\texttt{max\_tokens} for decision & \texttt{4} & \texttt{max\_tokens} for rationale & \texttt{120} \\
\midrule
\multicolumn{4}{c}{\textbf{Stage 2}}\\
\midrule
\textbf{Model Versions}: &  \\
\multicolumn{2}{l}{Gpt-4 API as generation backend} & \multicolumn{2}{l}{\texttt{gpt-4-turbo-2024-04-09}}\\
\textbf{Generation Configs}: &  \\
\texttt{temperature} & \texttt{1.0} & \texttt{candidates} & \texttt{1} \\
\midrule
\multicolumn{4}{c}{\textbf{Stage 3}}\\
\midrule
\multicolumn{2}{l}{\textbf{Base Model}:} & & \\
\multicolumn{2}{l}{LLaMA 3 8B} & \multicolumn{2}{l}{\texttt{meta-llama/Meta-Llama-3-8B}} \\
\multicolumn{2}{l}{Mixtral 8$\times$7B} & \multicolumn{2}{l}{\texttt{mistralai/Mixtral-8x7B-v0.1}} \\
\textbf{SFT}: & & \textbf{DPO}: &  \\
\texttt{finetuning\_type} & \texttt{lora} & \texttt{finetuning\_type} & \texttt{lora} \\
\texttt{lora\_target} & \texttt{all} & \texttt{lora\_target} & \texttt{all} \\
\texttt{num\_train\_epochs} & \texttt{4.0}& \texttt{num\_train\_epochs} & \texttt{3.0}\\
\texttt{batch\_size} & \texttt{8} & \texttt{batch\_size} & \texttt{1} \\
\texttt{gradient\_accumulation} & \texttt{8}& \texttt{gradient\_accumulation} & \texttt{8} \\
\texttt{lr\_scheduler\_type} & \texttt{cosine} & \texttt{lr\_scheduler\_type} & \texttt{cosine} \\
\texttt{learning\_rate} & \texttt{5e-5}& \texttt{learning\_rate} & \texttt{1e-5}\\
\texttt{quantization\_bit} & \texttt{4} & \texttt{quantization\_bit} & \texttt{4} \\
\texttt{fp16} & \texttt{true}& \texttt{fp16} & \texttt{true}\\
\texttt{upcast\_layernorm} & \texttt{true}& \texttt{upcast\_layernorm} & \texttt{true}\\
\texttt{warmup\_steps} & \texttt{0.1} & \texttt{warmup\_steps} & \texttt{0.1}\\
\textbf{Generation Configs}: & &  &  \\
\texttt{do\_samples} & \texttt{False} & \texttt{max\_new\_tokens}  & \texttt{256} \\
\midrule
\multicolumn{4}{c}{\textbf{Baselines and Score Extractor}}\\
\midrule
\textbf{Text Classifier}: & & \textbf{Score Extractor}: &  \\
\texttt{num\_train\_epochs} & \texttt{15.0}& \texttt{num\_train\_epochs} & \texttt{20.0}\\
\texttt{batch\_size} & \texttt{16}& \texttt{batch\_size} & \texttt{32}\\
\texttt{learning\_rate} & \texttt{2e-5}& \texttt{learning\_rate} & \texttt{1e-5}\\
\texttt{warmup\_steps} & \texttt{100}& \texttt{warmup\_steps} & \texttt{100}\\
\texttt{adam\_epsilon} & \texttt{1e-8}& \texttt{adam\_epsilon} & \texttt{1e-8}\\
\texttt{early\_stopping\_patience} &  \texttt{3} & \texttt{early\_stopping\_patience} &  \texttt{3} \\
\bottomrule
\end{tabular}
}
\caption{Hyper-parameters settings.}
\label{tab:hyper-parameters}
\end{table*}
\subsection{Model Implementation Details}
For our implementation, we utilized the HuggingFace Transformers package \cite{hf} for accessing pre-built language model architectures. In the third stage of our framework, the training and inference pipeline was set up using LLaMA-factory \cite{llama-factory}, along with the implementation of DPO \cite{dpo} and QLoRA \cite{qlora}. Due to limited computational resources, all of the training involved in the third stage uses 4-bit QLoRA.

For text classifiers baseline and the Score Extractor, we employed the \texttt{microsoft/deberta-v3-large} model \cite{debertav3}. We uses the LongT5 model \cite{longt5} for the AERA framework implementation. The \texttt{meta-llama/Meta-Llama-3-8B} \cite{llama3} and \texttt{mistralai/Mixtral-8x7B-v0.1} \cite{mixtral} were selected as the base models during the third stage of our framework.

Additionally, for accessing the GPT-4 \cite{gpt4} and Mixtral APIs, we referred to their respective official websites. 

We obey the licenses of all involved models. All of our models are trained with a single NVIDIA A100 80G GPU.
\subsection{Quadratic Weighted Kappa}
Quadratic Weighted Kappa, a widely used metric in evaluating the agreement between two raters in student response assessment, is defined as:
\begin{equation}
    \kappa = 1 - \frac{\sum_{i=1}^{k}\sum_{j=1}^{k} w_{ij} O_{ij}}{\sum_{i=1}^{k}\sum_{j=1}^{k} w_{ij} E_{ij}}
\end{equation}
where $k$ is the score set, $w$ is the weighted matrix, calculates as: $w_{i, j}=\frac{(i-j)^2}{(k-1)^2}$. $O$ is a $k\times k$ histogram matrix and $E$ being the $k\times k$ expected value matrix. 

\subsection{Score Extractor} \label{app:score_extractor}
Since our framework generates free-form rationales, the assessed mark could be embedded anywhere within the rationale context. We, therefore, train a Score Extractor utilizing a PLM-based text classifier on the rationale-to-score dataset constructed in Stage 2. The classifier is trained on 11,641 training data instances and validated with 2,319 instances, achieving a final accuracy of 1.0, a macro f1 score of 1.0, and a QWK score of 1.0. This indicates that our score extractor could accurately extract assessed marks from free-text rationales.

\subsection{Model Choice for AERA} \label{app:aera_model_choice}
We tried to align the model use and upgrade the rationale generation backend used in the AERA \cite{li-etal-2023-distilling} to GPT-4. However, we find that GPT-4 may sometimes fail on the label refinement step in the AERA framework. When utilising gpt-4-turbo-preview for rationale generation, an occasional occurrence arises where the model disregards the provided ground truth label hint and generates its own justification. E.g., ``\textit{0 points; 2.5 points; The student's answer begins to touch on several key elements but does not fully or clearly articulate them according to the criteria.}'' where ``0 points;'' are provided within the prompt, and the model supposed to continue to generate a rationale that justifies this decision.

We randomly sampled 89 instances and compared the rationale generated by GPT-4 and ChatGPT. We found 5 cases from GPT-4 that suffer from such an issue. This observation verifies our concern about the difficulty of LLMs solving complex reasoning tasks in one step. Interestingly, no single case from the ChatGPT output encountered deviations from the provided hint. 

\section{Framework Implementation Details}
In this section, we provide further details on the implementation and ablation studies we carried out for our framework.

\subsection{Stage 1 further details}
\label{app:stage1_further}

\subsubsection{Thought Tree Path Search Algorithm}
As demonstrated in Algorithm \ref{alg:search}, we adopt depth-first search in our implementation for both querying and traversing thought trees. We use Anytree\footnote{\url{https://pypi.org/project/anytree/}} to implement the tree node.
\begin{algorithm}
\caption{Tree Path Search Algorithm}
\begin{algorithmic}[1]
\Procedure{DFS}{parent, key\_elements}
    \State \textbf{Input:} Parent node and key elements list
    \State \textbf{Output:} Traversal of the thought tree
    \State
    \If{length(key\_elements)$ == 0$}
        \State \Return
    \EndIf
    \State
    \State decisions = \textit{Query}(key\_elements[0]) 
    \State \Comment{Obtain intermediate decisions}
    \State key\_elements.pop(0)
    \State \Comment{Remove queried key answer elements}
    \State \For{decision $\in$ decisions}
    \State new\_node = decision
    \State new\_node.parent = parent
    \State \Call{DFS}{new\_node, key\_elements}
    \EndFor
    \State \Comment{Assign new decision and traverse}
    \State 
\EndProcedure
\end{algorithmic}
\label{alg:search}
\end{algorithm}
\subsubsection{Thought Tree Prompt Design}
As shown in Figure \ref{tcb:universal_thought_tree}, we provide a universal prompt template we used to query LLMs.
\begin{tcolorbox}[title=Prompt Templates for Thought Trees,
    colback=white,
    colframe=yellow!75!black,
    colbacktitle=yellow,
    coltitle=black,
    breakable,
    label=tcb:universal_thought_tree,
    fonttitle=\bfseries]
\textbf{Query decision for $k_j$}:\\
\text{[Key Answer Element]}: Does this student answer specify \text{[}\textit{fill in the current key answer element content}\text{]}? Please select Yes for matching this key answer element and No for non-matching this key answer element.\\
\{\{demonstation\}\}\\
\text{[Student Answer]}: "\{\{student\_answer\}\}"\\
\text{[Decision]}: \\
\textbf{Query rationale $r_j$ for $k_j$}:\\
Please generate a rationale that justifies your above decision. Quote ** part of this student answer ** (e.g. just a few words) that answers this key element with "...". Do not include any other additional information that are not relevant to this decision.
\end{tcolorbox}

\subsubsection{Marking Rubric to Python Code}
As shown in Figure \ref{tcb:marking_rubric_python}, we present an example Python code transformed by the GPT-4 model for ASAP 1's marking rubrics. Our framework is implemented to automatically use a Python interpreter to execute this code and dynamically used to summarize marks from a list of key answer element decisions. 
\begin{tcolorbox}[title=Marking Rubric to Python Code,
    colback=white,
    colframe=yellow!75!black,
    colbacktitle=yellow,
    coltitle=black,
    breakable,
    label=tcb:marking_rubric_python,
    fonttitle=\bfseries]
\textbf{Original Marking Rubric}\\
3 points: The response describes three additional pieces of information that would be needed to accurately replicate the experiment;\\
2 points: The response describes two additional pieces of information that would be needed to accurately replicate the experiment;\\
1 point: The response describes one additional piece of information that would be needed to accurately replicate the experiment;\\
0 point: The response describes little or no accurate or relevant information from the acid rain investigation.\\
\textbf{GPT-4 Transformed Python Code}:\\
def sum\_score(node\_names):\\
\phantom{1234}num\_of\_correct="; ".join(node\_names).count("Yes")\\
\phantom{1234}if num\_of\_correct >= 3:\\
\phantom{1234}\phantom{1234}return 3\\
\phantom{1234}elif num\_of\_correct == 2:\\
\phantom{1234}\phantom{1234}return 2\\
\phantom{1234}elif num\_of\_correct == 1:\\
\phantom{1234}\phantom{1234}return 1\\
\phantom{1234}else:\\
\phantom{1234}\phantom{1234}return 0
\end{tcolorbox}

\subsubsection{Ablation Study on Prompt Structure}
We invested significant effort to identify the optimal prompt structure for querying thought trees. This section outlines our rationale behind the final template structure used in stage 1 of our framework.

\begin{figure*}[t]
\centering
\includegraphics[width=\linewidth]{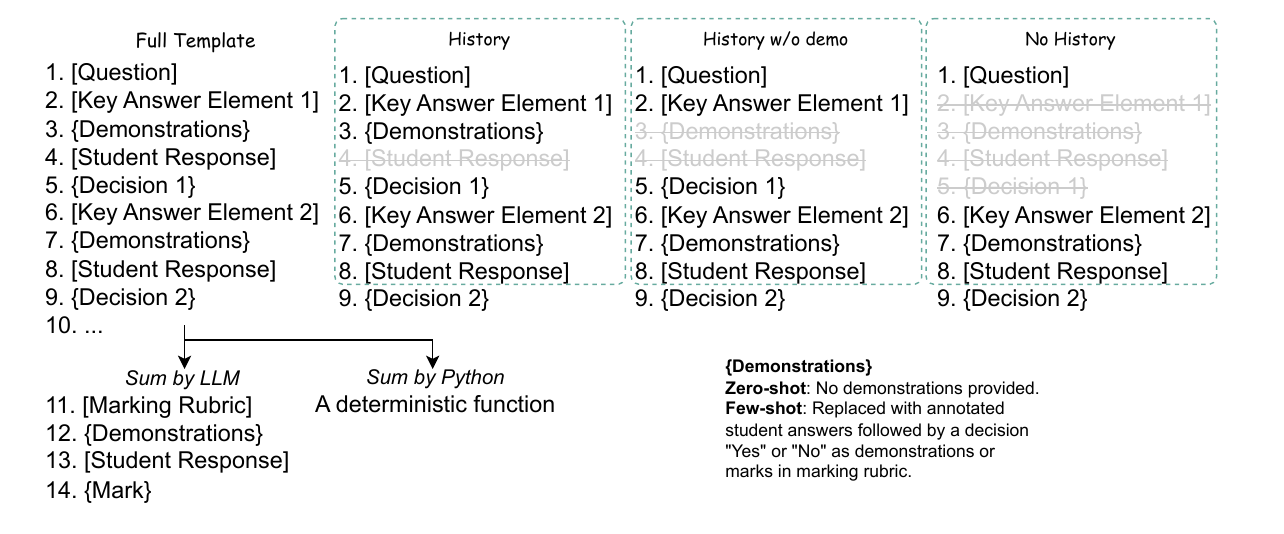}
\caption{Explanation of our prompt structure using querying decision 2 as an example.}
\label{fig:prompt_structure}
\end{figure*}

\textbf{TL;DR}: Few-shot prompt template in ``History w/o demos'' mode proved most effective.

\textbf{Prompt Template Setup}: We sampled several hundred instances from the Private 1 dataset's training set for this trial experiment, as it only contains four key answer elements, the fewest among all our datasets. As depicted in Figure \ref{fig:prompt_structure}, our prompt template included two demonstration setups: zero-shot and few-shot. In the zero-shot mode, no demonstrations are provided. In the few-shot mode, each score level includes an annotated student answer, with the annotated example key answer element decisions labelled as ``Yes'' (if the answer answered the key element) or ``No'' (if not). At the marking rubric assessment stage, we designed two marking rubric application modes in our framework: ``Sum by Python'', which uses a deterministic Python function, and ``Sum by LLM'', which intuitively provides the text descriptive marking rubrics to the LLM and query LLMs to summarize a score. However, we didn't use ``Sum by LLM'' in our final thought tree for reasons detailed in the subsequent experimental analysis. 

As shown in Figure \ref{fig:prompt_structure}, we designed three history modes: ``History'', which includes full history from previous decisions (e.g., when querying decision 2, we retain key answer element 1 and its demonstrations along with decision 1 query result); ``History w/o demos'', which omits demonstrations used in the last session but retains only the decisions (e.g., when querying decision 2, we retain key answer element 1 and decision 1 query result, but removed the demonstration from the history); and ``No History'', treating each key answer element as a separate, parallel query (e.g., the prompt only contains the current key element's demonstration).

\begin{table*}[]
\resizebox{\linewidth}{!}{%
\begin{tabular}{c|cccccc}
\multicolumn{1}{c|}{\textbf{No}} & \multicolumn{1}{c}{\textbf{Prompt Mode}} & \multicolumn{1}{c}{\textbf{w/wo history}} & \multicolumn{1}{c}{\textbf{Marking Rubric Format}} & \multicolumn{1}{c}{\textbf{Acc}} & \multicolumn{1}{c}{\textbf{F1}} & \multicolumn{1}{c}{\textbf{QWK}} \\ \hline
1                                & Zero-shot                                         & no history                                & Python                                    & 0.4173                                 & 0.3760                                 & 0.6459                           \\
2                                & Zero-shot                                         & no history                                & LLM                                       & 0.4252                                 & 0.3876                                 & 0.6481                           \\
3                                & Zero-shot                                         & history w/o demo                           & Python                                    & 0.5039                                 & 0.4782                                 & 0.7355                           \\
4                                & Zero-shot                                         & history w/o demo                           & LLM                                       & 0.5157                                 & 0.4910                                 & 0.7491                           \\
5                                & Few-shot                                          & no history                                & Python                                    & 0.4803                                 & 0.4474                                 & 0.7370                           \\
6                                & Few-shot                                          & no history                                & LLM                                       & 0.4843                                 & 0.4391                                 & 0.7432                           \\
7                                & Few-shot                                          & history                                   & Python                                    & 0.4528                                 & 0.4157                                 & 0.7278                           \\
8                                & Few-shot                                          & history                                   & LLM                                       & 0.4685                                 & 0.4444                                 & 0.7544                           \\
9                                & Few-shot                                          & history w/o demo                           & Python                                    & \textbf{0.5354}                                 & \textbf{0.5116}                                 & \textbf{0.7791}                           \\
10                               & Few-shot                                          & history w/o demo                           & LLM                                       & 0.5354                                 & 0.5098                                 & 0.7690                           \\ \hline
11                               & Few-shot (Duo)                                    & no history                                & Python                                    & 0.4567                                 & 0.4139                                 & 0.7103                           \\
12                               & Few-shot (Duo)                                    & no history                                & LLM                                       & 0.4961                                 & 0.4608                                 & 0.7287                           \\
13                               & Few-shot (Duo)                                    & history                                   & Python                                    & 0.4685                                 & 0.4380                                 & 0.7397                           \\
14                               & Few-shot (Duo)                                    & history                                   & LLM                                       & 0.4724                                 & 0.4400                                 & 0.7528                           \\
15                               & Few-shot (Duo)                                    & history w/o demo                           & Python                                    & 0.5197                                 & 0.4977                                 & 0.7721                           \\
16                               & Few-shot (Duo)                                    & history w/o demo                           & LLM                                       & 0.5236                                 & 0.4991                                 & 0.7537                           \\ \hline
17                               & Few-shot (Shuffle)                                & history                                   & Python                                    & 0.4528                                 & 0.4186                                 & 0.7182                           \\
18                               & Few-shot (Shuffle)                                & history                                   & LLM                                       & 0.5039                                 & 0.4866                                 & 0.7551                           \\
19                               & Few-shot (Shuffle)                                & history w/o demo                           & Python                                    & 0.5079                                 & 0.4693                                 & 0.7533                           \\
20                               & Few-shot (Shuffle)                                & history w/o demo                           & LLM                                       & 0.5433                                 & 0.5219                                 & 0.7804                           \\ \hline
21                               & Few-shot (Duo) (Shuffle)                          & history                                   & Python                                    & 0.4921                                 & 0.4678                                 & 0.7062                           \\
22                               & Few-shot (Duo) (Shuffle)                          & history                                   & LLM                                       & 0.4921                                 & 0.4615                                 & 0.7173                           \\
23                               & Few-shot (Duo) (Shuffle)                          & history w/o demo                           & Python                                    & 0.5394                                 & 0.5272                                 & 0.7614                           \\
24                               & Few-shot (Duo) (Shuffle)                          & history w/o demo                           & LLM                                       & 0.5512                                 & 0.5328                                 & 0.7879                          
\end{tabular}}
\caption{Thought Tree Assessment Performance with Different Prompt Structure on GPT-4.}
\label{tab:prompt_setups}
\end{table*}

\textbf{Demonstration Examples Setup}: ``Few-shot'' contains 5 examples picked from scores ranging from 0-4. Each instance has been annotated to whether contains key element $k_j$ or not. ``Few-shot (Duo)'' contains 2 examples. We picked a full-score example in contrast to a zero-score example. ``Shuffle'': For ``history'' and ``history w/o demo'' mode, each current key answer element decision could be influenced by the previous key answer element provided in the history. Therefore, we carried out an ablation study by shuffling the order of input key answer elements to examine the influence of input order.

\textbf{Experimental Results}: We present a comprehensive experiment analysis in Table \ref{tab:prompt_setups}. Our analysis reveals that the scoring for different key elements is hierarchical, influenced by the sequential processing of elements and the demonstration provided. The use of few-shot examples, especially in the ``History w/o demos'' mode, significantly enhanced the accuracy of key element decisions, as evidenced by the improved performance in setups 3 Vs. 9 and 1 Vs. 5. History including demonstrations (7 Vs. 9) did not prove beneficial, this may due to the reason that previous key elements' demonstrations may influence the effect of demonstrations for the current key answer element and confuses the LLM to make confident decisions.

Since the key element-level decision only requires ``Yes'' or ``No'', we present (Duo) demonstrations, which means we only provide a pair of demonstrations that contain (``Yes'') and not contain (``No'') the key answer element. By comparing five demonstration examples with two demonstration examples (5 Vs. 11, 9 Vs. 15), we find that providing more demonstration examples can result in better results.

We also investigated the influence of order on the key answer elements, denoted as (Shuffle) in the table. We find that different order settings of key answer elements could result in fluctuation in assessment performance. However, when human assessors assess the student responses, they most likely follow a top-to-bottom order. Therefore, we keep the original key answer element orders for each question.

\begin{figure}[h]
    \centering
    \begin{minipage}{0.5\linewidth}
        \centering
        \includegraphics[width=1.0\linewidth]{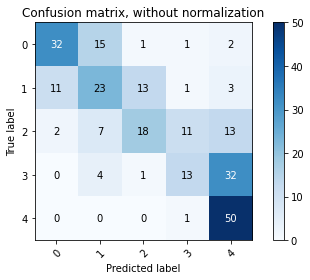} %
    \end{minipage}\hfill
    \begin{minipage}{0.5\linewidth}
        \centering
        \includegraphics[width=1.0\linewidth]{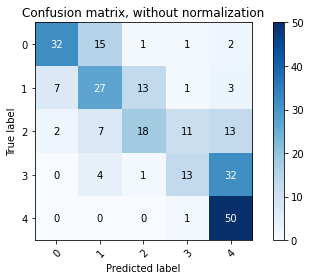} %
    \end{minipage}
    \caption{Comparison of confusion matrix for ``Sum by Python'' (left) and ``Sum by LLM'' (right).}
    \label{fig:python_llm_matrix}
\end{figure}

In most cases, the ``Sum by LLM'' method improved over the ``Sum by Python'' approach. We illustrated this comparison in Figure \ref{fig:python_llm_matrix}, where the confusion matrices for experiments 9 and 10 highlight the differences between the two methods. While ``Sum by LLM'' appears to enhance results, it does not adhere as strictly to the original assessment tree as the ``Sum by Python'' method, which is purely deterministic. Interestingly, ``Sum by LLM' seems to ignore previous key answer element-level assessment decisions and provide a final mark that potentially gives higher scoring performance. This unpredictable improvement led us to exclude ``Sum by LLM'' as the source of generating synthetic rationales. However, this approach holds potential for future investigation, particularly if combined with strategies like performing self-reflection to correct erroneous intermediate decisions thought paths.
\subsubsection{Backend Model Choices} \label{app:model_choice}
As illustrated in Table \ref{tab:model_choices}, we comprehensively evaluated various model options to query thought trees and imitate the real human assessment processes. We discovered that only GPT-4 and Mixtral 8$\times$7B could accurately capture and reason based on our prompt design.

We find that the LLaMA 2 series models \cite{llama2} struggled with adhering to the prompt requirements. Although the model \texttt{google/gemma-7b-it} \cite{gemma} was able to follow the prompt, it often incorporated excessive context. In contrast, google/gemma-2b-it correctly adhered to the prompt and made appropriate judgments. However, before each tree query, we implemented a preliminary check of basic computational accuracy—querying the model with ``what is $6+7=$''—revealed that \texttt{google/gemma-2b-it} was the only model to incorrectly return $47$ as the answer, indicating potential issues with its reliability for our purposes.

\begin{table*}[!h]
\centering
\resizebox{\linewidth}{!}{
\begin{tabular}{|p{\linewidth}|}
\hline
\textbf{Prompt}: \\
\textbf{[Question]}: Microorganisms in the soil recycle carbon from the leaves so that the carbon is used for new plant growth. Explain how. [4 marks] \\
\textbf{[Key Answer Element]}: Does this answer address the decay or decomposition of leaves? Additionally, consider if the answer includes aspects of how leaves are broken down. Please select \textit{Yes} for matching this key answer element and \textit{No} for non-matching this key answer element.\\
\textbf{[Student Answer]}: "[\textit{An example student answer}]" \textbf{[Decision]}: Yes \\
\textbf{[Student Answer]}: "[\textit{An example student answer}]"  \textbf{[Decision]}: Yes \\
\textbf{[Student Answer]}: "[\textit{An example student answer}]"  \textbf{[Decision]}: Yes \\
\textbf{[Student Answer]}: "[\textit{An example student answer}]"  \textbf{[Decision]}: No \\
\textbf{[Student Answer]}: "[\textit{An example student answer}]"  \textbf{[Decision]}: No \\
\textbf{[Student Answer]}: "[\textit{A student answer needed to be assessed}]"  \textbf{[Decision]}: \\
\hline
\textbf{mistralai/Mixtral-8x7B-Instruct-v0.1}: \\
No, this answer discusses the movement of carbon through the food chain, not the decomposition of leaves in the soil. \\
\textbf{meta-llama/Llama-2-7b-chat-hf}: \\
Great, it looks like you're on the right track! Here's how you can improve your answer:\\
1. You're partially correct when you say that microorganisms in the soil recycle carbon from leaves. However, you need to provide more detail on how this process works. For example, you could explain that microorganisms such as bacteria and fungi break down the leaves into simpler compounds, releasing carbon dioxide into the atmosphere through respiration. This carbon dioxide is then absorbed by plants during photosynthesis, which produces glucose that contributes to new plant growth.\\
\textbf{meta-llama/Llama-2-13b-chat-hf}: \\
Here are the decisions for each student answer:\\
1. [Student Answer]: "the example student answer"\\
Decision: Yes\\
This answer addresses the decay or decomposition of leaves and includes aspects of how leaves are broken down. It correctly explains that microorganisms decompose leaves, releasing carbon dioxide into the atmosphere, which is then taken in by plants during photosynthesis to produce glucose for growth.\\
\textbf{meta-llama/Llama-2-70b-chat-hf}\\
Yes, all of the student answers address the decay or decomposition of leaves and include aspects of how leaves are broken down.\\
The first student answer explains that microorganisms decompose dead leaves, releasing carbon dioxide into the atmosphere, which is then taken in by plants during photosynthesis and used for growth.\\
\textbf{google/gemma-7b-it}\\
\# Analysis of Student Answers\\
**The question:** Microorganisms in the soil recycle carbon from the leaves so that the carbon is used for new plant growth.\\
**Key Answer Element:** Does this answer address the decay or decomposition of leaves? Additionally, consider if the answer includes aspects of how leaves are broken down.\\
**Student Answer 1:**\\
* **Decision:** Yes\\
\textbf{google/gemma-2b-it}\\
No\\
\hline
\end{tabular}}
\caption{Comparison output with different open source LLMs.}
\label{tab:model_choices}
\end{table*}

\subsubsection{Assessment Performance for Thought Tree with Different Backends}
As highlighted in \textsection{\ref{app:model_choice}}, only GPT-4 and Mixtral 8$\times$7B could effectively handle our designed prompts. In this section, we compare the assessment thought tree's performance of these two models. We consider paths with the highest predicted probability as the predicted mark.

Table \ref{tab:tree_performance} outlines the performance metrics on test sets for both backends. Evidently, GPT-4 significantly outperforms Mixtral on the Private 1 and ASAP datasets 1, 2, and 5. We also calculated the correct ratio:
\begin{equation}
\text{correct ratio} = 
\frac{\text{Num of tree containing correct mark}}{\text{Total num of trees}} \notag
\end{equation}
This correct ratio calculates how many tree paths contain the ground-truth labels, although it might not be the highest probability path. As shown in Table \ref{tab:tree_performance}, counting other lower predictive probability paths could potentially increase the accuracy near 95\%, with an average improvement from 14\% to 27\%. This insight is crucial as it implies a higher success rate for synthetic data generation in Stage 2.

\begin{table}[!h]
\centering
\resizebox{\linewidth}{!}{%
\begin{tabular}{lcccc|ccc}
\toprule
Backend & \multicolumn{4}{c}{GPT-4} & \multicolumn{3}{c}{Mixtral 8$\times$7B} \\ 
\cmidrule(r){2-5} \cmidrule(l){6-8} 
\textbf{Datasets} & Acc & F1 & \textbf{QWK} & Correct Ratio & Acc & F1 & \textbf{QWK} \\
\midrule
Private 1 & 0.5276 & 0.5078 & \textbf{0.7891} & 0.6614 & 0.3780 & 0.2996 & 0.6526 \\
Private 2 & 0.5459 & 0.5362 & \textbf{0.6710} & 0.7806 & 0.4643 & 0.4454 & 0.6122 \\
ASAP 1    & 0.6011 & 0.6145 & \textbf{0.7419} & 0.8610 & 0.4386 & 0.4351 & 0.6285 \\
ASAP 2    & 0.6746 & 0.6852 & \textbf{0.7856} & 0.9437 & 0.2911 & 0.2793 & 0.3469 \\
ASAP 5    & 0.8043 & 0.5317 & \textbf{0.7834} & 0.9482 & 0.7102 & 0.3999 & 0.6750 \\
ASAP 6    & 0.7496 & 0.4429 & 0.5831 & 0.9299 & 0.8446 & 0.5060 & \textbf{0.6716} \\
\bottomrule
\end{tabular}
}
\caption{Thought Tree Assessment Performance with Different Backend Options.}
\label{tab:tree_performance}
\end{table}

\subsubsection{Cost Analysis}
Table \ref{tab:gpt4_cost} details the cost analysis for obtaining thought trees with GPT-4 as backend. The reported cost includes the total query cost for train, validation and test sets. The cost for querying each question highly correlates to the length of question information and the size of the key answer elements set. Our framework will take more cost in the query if the question contains more key answer elements.
\begin{table}[!h]
\centering
\begin{tabular}{cc}
\toprule
\textbf{Datasets} & Cost (in USD) \\
\midrule
Private 1 & 215.96\\
Private 2 & 181.25 \\
ASAP 1    & 2,116.28 \\
ASAP 2    & 705.21 \\
ASAP 5    & 913.23 \\
ASAP 6    & 931.28 \\
\bottomrule
\end{tabular}
\caption{API Cost Analysis for Querying.}
\label{tab:gpt4_cost}
\end{table}

\subsection{Stage 2 Further Details}
\label{app:stage2_further}

\subsubsection{Prompt Template for Rationale Summarization}
We used the prompt template provided in Figure \ref{tcb:summarization_prompt} to summarize response-level rationales and improvement suggestions in json format \cite{li_json_format}, to obtain a free-form textual output.

\begin{tcolorbox}[title=Prompt Template for Response-Level Rationale Summarization for GPT-4,
    colback=white,
    colframe=yellow!75!black,
    colbacktitle=yellow,
    coltitle=black,
    breakable,
    label=tcb:summarization_prompt,
    fonttitle=\bfseries]
Here is a student answer to the following question:\\
"\{question\}"\\
\text{[Student Answer]}: "\{student\_answer\}"\\
This question follows a points mark scheme, and the breakdown assessment by each Key Answer Element for this student's answer is as follows:\\
\{idx\}. \{key\_element\} - \{decision\}: \{rationale\}\\
...\\
\\
According to the: "\{rubric\}", the answer should get a score of \{tree\_predicted\_score\}.\\
\\
Please summarize the above rationales and be FAITHFUL to the given assessment decisions for this student's answer briefly and precisely. Give the summarization in JSON format:\\
\text{\char0{}\char0{}\char0{}JSON}\\
\{\\
\phantom{1234}"mark": "...", \# numeric\\
\phantom{1234}"rationale": "...", \# including mark awarded, which marking rubric applied, and detailed key elements level rationale\\
\phantom{1234}"suggestion": "..." \# any answer improvement suggestion\\
\}\text{\char0{}\char0{}\char0{}}\\
- The "mark" should be the score of the student's answer.\\
- The "rationale" should be concise, include the assessed score and rubric applied, more importantly, justify the **marking decision-making processes** by **summarizing** the key element-level rationales, you must **quote the exact part** from the student's answer.\\
- If the student didn't get a full mark, you can also provide some suggestions for improvement; otherwise, leave it blank.
\end{tcolorbox}
\subsubsection{Synthetic Generated Data Statistics}
\begin{figure}[t]
  \centering
  \includegraphics[width=\linewidth]{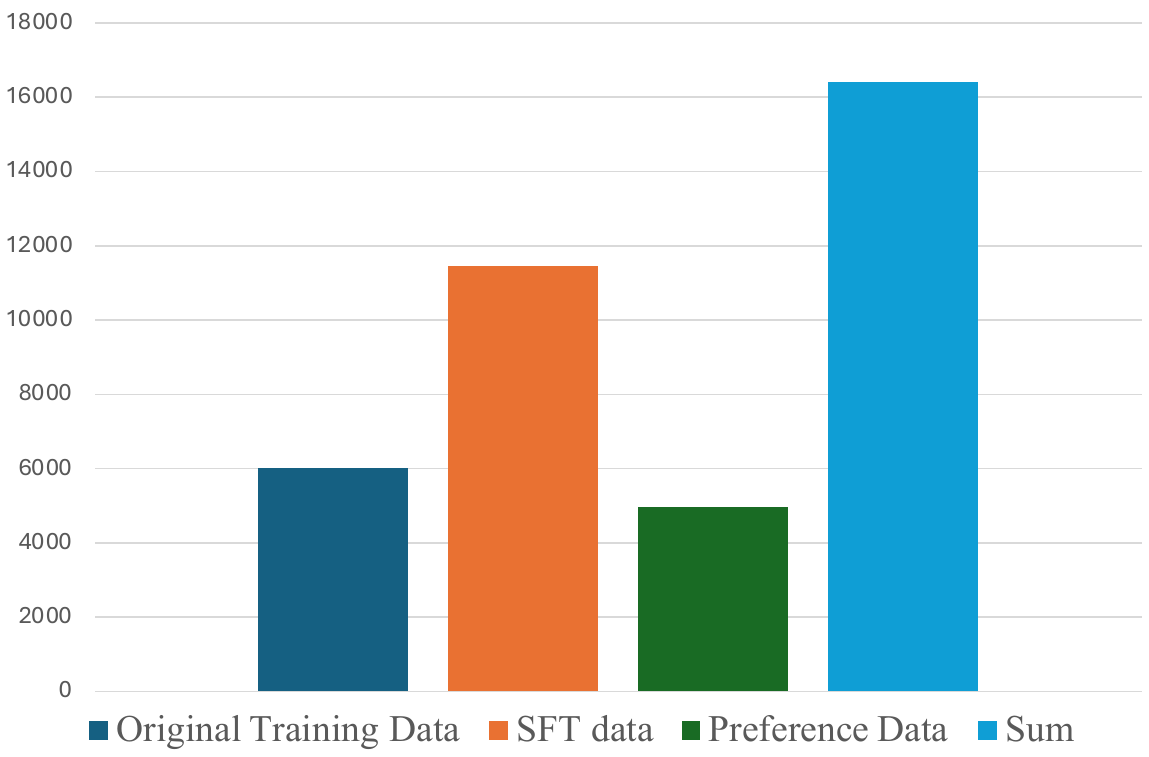}
  \caption{Data statistics of generated synthetic data.}
  \label{fig:data_stats}
\end{figure}

This section presents statistics for the synthetic SFT and preference data generated across three settings: all datasets, all datasets excluding ASAP 2, and all datasets excluding our Private data. Table \ref{tab:synthetic_data_statistic} details the distribution of this data across training and validation subsets.
\begin{table}[!h]
\centering
\begin{tabular}{lcc}
\toprule
\textbf{Datasets} & Train & Validation \\
\midrule
Rationale to score & 11,641 & 2,319 \\
\midrule
SFT all datasets & 11,450   & 2,771  \\
SFT w/o ASAP 2 & 9,607   & 2,314  \\
SFT w/o Private & 10,126   & 2,520  \\
\midrule
DPO all datasets & 4,960   & 705  \\
DPO w/o ASAP 2 & 4,635  & 624  \\
DPO w/o Private & 4,799   & 676  \\
\bottomrule
\end{tabular}
\caption{Statistics for Generated Synthetic Rationale Data.}
\label{tab:synthetic_data_statistic}
\end{table}
A primary motivation for this study is to address data scarcity issues inherent in the student answer scoring datasets in general. As illustrated in Figure \ref{fig:data_stats}, our method generated more than three times the amount of synthetic data compared to the original dataset, significantly enriching the available data resources for training.
\subsubsection{Prompt Instructions for the SFT Data}
In this section, we include the instruction prompt for rationale generation and generate improvement suggestions, which are randomly assigned to each instance.
\paragraph{Rationale Instructions}
\begin{itemize}
\setlength{\itemsep}{1pt}
  \setlength{\parskip}{0pt}
  \setlength{\parsep}{0pt}
    \item ``Evaluate the [Student Answer] based on the provided [Key Answer Elements] and [Marking Rubric]. Summarize the assessment and justify the score awarded: ''
    \item ``Analyze the [Student Answer] thoroughly. Generate a detailed rationale that explains the strengths and weaknesses of the response:''
    \item ``Conduct a comprehensive evaluation of the [Student Answer] using the provided [Key Answer Elements] and any relevant background information. Provide a detailed rationale for the assessment, including specific references to the [Marking Rubric]:''
    \item ``Critically assess the [Student Answer] in light of the [Key Answer Elements]. Offer a detailed explanation for the score assigned, referencing specific criteria from the [Marking Rubric]:''
    \item ``Review the [Student Answer] for accuracy and completeness compared to the [Key Answer Elements]. Provide a comprehensive rationale that includes the strengths, weaknesses, and areas for improvement:''
    \item ``Examine the [Student Answer] thoroughly, comparing it against the [Key Answer Elements] and using the guidelines provided in the [Marking Rubric]. Clearly articulate the reasoning behind each point awarded or deducted:''
    \item ``Systematically evaluate the [Student Answer] by aligning it with the [Key Answer Elements] and assessing according to the [Marking Rubric]. Document your findings and the logic of your evaluation comprehensively:''
\end{itemize}
\paragraph{Suggestion Instructions}
\begin{itemize}
\setlength{\itemsep}{1pt}
  \setlength{\parskip}{0pt}
  \setlength{\parsep}{0pt}
    \item ``Analyze the [Student Answer] and identify areas for improvement. Suggest specific changes that could enhance the clarity, accuracy, and depth of the response:''
    \item ``Please provide targeted suggestions to help the student refine their answer for better alignment with the expected criteria:''
    \item ``Please offer some actionable feedback that the student can use to improve their response:''
    \item ``Could you propose concrete improvements that could elevate the response’s overall quality and effectiveness:''
    \item ``Please recommend enhancements that can strengthen the logic, persuasiveness, and completeness of the answer:''
\end{itemize}

\section{Further Experimental Results}
\label{app:further_experiments}

\subsection{Comprehensive Analysis on Framework Effectiveness}
As presented in Table \ref{tab:further_assessment_results}, we provide a comprehensive comparison of the effectiveness of our framework on two base models: Mixtral 8$\times$7B and LLaMA 3 8B, with three dataset settings:
\begin{itemize}
\setlength{\itemsep}{1pt}
  \setlength{\parskip}{0pt}
  \setlength{\parsep}{0pt}
    \item[] \textbf{Trained on All Data}: Models are trained on all the SFT \& DPO datasets.
    \item[] \textbf{Trained on All Data w/o ASAP2}: Models are trained on SFT \& DPO w/o ASAP 2 datasets.
    \item[] \textbf{Trained on All Data w/o Private data}: Models are rained on SFT \& DPO w/o Private datasets.
\end{itemize}
Table \ref{tab:synthetic_data_statistic} presents detailed data statistics for each dataset combination.

\paragraph{Experiments with LLaMA 3 8B} 
As demonstrated in Table \ref{tab:further_assessment_results}, we extended our experiments to LLaMA 3 8B model. When compared with AERA's results from Table \ref{tab:model_comparison}, the SFT model trained on LLaMA exhibited an $11.17\%$ performance improvement. Consistent with observations in \textsection{\ref{sec:overall_assessment}}, our results on LLaMA further confirm that preference modelling (DPO training step) can enhance assessment performance by $2\%$ compared to SFT. Despite a minor decrease in the QWK score on the ASAP 5 dataset, both accuracy and F1 scores showed improvements. These results demonstrate that our framework works not only with a mixture of expert models, reinforcing its robustness and adaptability.

\paragraph{Experiments on Mixtral 8$\times$7B} 
As demonstrated in Table \ref{tab:further_assessment_results}, our experiments for Mixtral 8$\times$7B on three dataset settings show superiority over the LLaMA 3 8B as a base model choice. This is probably due to Mixtral 8$\times$7B has shown advanced reasoning capability compared with LLaMA 3 8B \cite{mixtral}. However, compared with the size of the parameters and the test performance, LLaMA 3 8B is still a competitive model choice considering the training cost.

\paragraph{Effectiveness on DPO}
As demonstrated in Table \ref{tab:further_assessment_results}, our six experiment combinations (two base models $\times$ three datasets settings) show the effectiveness of rationale preference optimization on the generated synthetic preference data. Interestingly, experiments show that models tested on unseen datasets may bring negative effects. However, this effect is possibly influenced by the similarity of the questions. For example, the Private 1 question shares a similar format to the ASAP 5 \& 6, which probably caused the performance actually to improve on the ``w/o Private data'' setting.

\paragraph{Transferability on Unseen Datasets}
Given that ASAP 2 is recognized as the most challenging dataset in \cite{li-etal-2023-distilling}, we conducted an ablation study by excluding ASAP 2 from the training set to evaluate the model's transferability on unseen data. Table \ref{tab:further_assessment_results} shows performance metrics for both SFT and DPO stages trained with Mixtral. Despite removing ASAP 2 from training, all other datasets improved after the DPO training step, albeit with a slight drop in QWK for ASAP 2. This is quite intuitive, as ASAP 2 hasn't been trained during any phases, so the preference for ASAP 2 hasn't been properly modelled in the dataset. By comparing with Table \ref{tab:model_comparison}, we can still see a performance improvement than the AERA's result in both SFT ($7.63\%$) and DPO ($10.79\%$) stages. This demonstrates the assessment ability could go beyond its training data.

\begin{table}[!h]
\centering
\resizebox{\linewidth}{!}{
\begin{tabular}{p{\linewidth}}
\small
\textbf{Student response}: Based on the students data I can \textcolor{red}{conclude that plastic type B has the largest amount stretched}. Two ways that the student could have improved the experimental decison is by allowing the weights to hang longer to see if the plastic would stretch more. Another way he could have improved his experiment is by \textcolor{red}{doing a third trials to really be sure that his results are correct}.\\
\small
\textbf{SFT w/ all data}: The student's answer \textcolor{red}{scored 2 points} according to the marking rubric. The student \textcolor{red}{correctly drew a valid conclusion, stating that 'plastic type B has the largest amount stretched,'} which aligns with the data provided. Additionally, the student \textcolor{red}{proposed one valid experimental design improvement by suggesting to 'do a third trials to really be sure that his results are correct,'} addressing the need for additional trials to confirm results. However, the student did not address other potential improvements such as ensuring uniform sample sizes, thickness, or the consistency in the weight added, which are crucial for experimental validity.\\
\small
\textbf{SFT w/o ASAP 2}: The student's answer \textcolor{red}{scored 2 points} according to the marking rubric, as it \textcolor{red}{'draws a valid conclusion supported by the student's data and describes one way the student could have improved the experimental design and\/or the validity of the results.'} The student correctly identified that \textcolor{red}{'plastic type B has the largest amount stretched,'} which is supported by the data. However, the student failed to address other key elements such as the comparative stretchability of other plastics, variations in thickness, and the exact number of weights used. The student \textcolor{red}{did suggest improvements like 'allowing the weights to hang longer' and 'doing a third trials,'} which aligns with the rubric's requirement for describing experimental design improvements. \\
\end{tabular}}
\caption{Comparison of Generated Rationale between SFT trained with All Data and All Data w/o ASAP 2.}
\label{tab:asap2_case_study}
\end{table}

Interestingly, as detailed in the case study presented in Table \ref{tab:asap2_case_study}, no significant quality differences are observed between assessment rationales generated by SFT Mixtral models trained with or without ASAP 2. Both rationales accurately capture the student's valid conclusion and proposed experimental improvements according to the marking rubric. They all agreed that another improvement suggestion was not valid, directly quoting the original phrases from the student's response.

Moreover, we can continue to observe this trend by observing the experiments trained with ``w/o Private data''. Despite the failure of LLaMA 3 8B on the Private 2 dataset, all the other results still maintain a comparable QWK score, as the results were trained with ``all data''. 

\paragraph{Ablation Study on Decoding Method}
We performed an ablation study on the decoding strategy as shown in the last two rows from Table \ref{tab:further_assessment_results}. We set the sample as True to use beam search on both LLaMA 3 and Mixtral 8$\times$7B SFT results. Interestingly, overall assessment performance dropped by 4.75\% and 3.62\% for LLaMA and Mixtral models, respectively. Therefore, we chose to use greedy decoding throughout our experiments.

\paragraph{Performance on Benchmark Datasets}
We also evaluated our tuned model after the DPO stage on general reasoning capability benchmarks using the \texttt{lm-evaluation-harness} \cite{eval-harness}. As shown in Table \ref{tab:benchmark_capability},
our evaluation demonstrates that the task-specific LLaMA 3 8B DPO model maintains robust reasoning capabilities across various benchmarks. Notably, it shows improved performance in commonsense reasoning and question-answering benchmarks such as BoolQ \cite{clark-etal-2019-boolq}, HellaSwag \cite{zellers-etal-2019-hellaswag}, OpenBookQA \cite{mihaylov-etal-2018-suit}, and PIQA \cite{bisk2019piqareasoningphysicalcommonsense}, despite some decreases in performance on other benchmarks like ARC Challenge and ARC Easy \cite{allenai:arc}. This suggests a targeted improvement in educational assessment rationale generation didn't degrade general reasoning capability for LLMs.

\begin{table*}[h]
\centering
\resizebox{0.9\textwidth}{!}{
\begin{tabular}{@{}lcccc@{}}
\toprule
\textbf{Datasets} & \textbf{Metric} & \textbf{LLaMA 3 8B} & \textbf{LLaMA 3 8B DPO (Ours)} & \textbf{Difference} \\ \midrule
ARC Challenge    & acc    & $0.4625 \pm 0.0146$ & $0.4471 \pm 0.0145$ & $-0.0154$ \\
\cite{allenai:arc}& acc\_norm & $0.5000 \pm 0.0146$ & $0.4812 \pm 0.0146$ & $-0.0188$ \\
ARC Easy         & acc    & $0.7832 \pm 0.0085$ & $0.7576 \pm 0.0088$ & $-0.0256$ \\
\cite{allenai:arc} & acc\_norm & $0.7521 \pm 0.0089$ & $0.7024 \pm 0.0094$ & $-0.0497$ \\
BoolQ \cite{clark-etal-2019-boolq} & acc    & $0.8015 \pm 0.0070$ & $0.8330 \pm 0.0065$ & $+0.0315$ \\
HellaSwag         & acc    & $0.5873 \pm 0.0049$ & $0.5974 \pm 0.0049$ & $+0.0101$ \\
\cite{zellers-etal-2019-hellaswag}         & acc\_norm & $0.7799 \pm 0.0041$ & $0.7862 \pm 0.0041$ & $+0.0063$ \\
OpenBookQA        & acc    & $0.3440 \pm 0.0213$ & $0.3540 \pm 0.0214$ & $+0.0100$ \\
\cite{mihaylov-etal-2018-suit}   & acc\_norm & $0.4340 \pm 0.0222$ & $0.4460 \pm 0.0223$ & $+0.0120$ \\
PIQA              & acc    & $0.7867 \pm 0.0096$ & $0.7949 \pm 0.0094$ & $+0.0082$ \\
\cite{bisk2019piqareasoningphysicalcommonsense}       & acc\_norm & $0.7949 \pm 0.0094$ & $0.8003 \pm 0.0093$ & $+0.0054$ \\
WinoGrande \cite{sakaguchi2019winogrande} & acc    & $0.7364 \pm 0.0124$ & $0.7277 \pm 0.0125$ & $-0.0087$ \\
\bottomrule
\end{tabular}}
\caption{Performance Comparison between LLaMA 3 8B (4-bit) and LLaMA 3 8B DPO (4-bit LoRA) Models on Various General Reasoning Capability Benchmarks.}
\label{tab:benchmark_capability}
\end{table*}

\begin{sidewaystable*}
    \centering
\resizebox{\textwidth}{!}{
\begin{tabular}{ccccccccccccccccccccccc}
\multicolumn{2}{c|}{Test   Data}                                    & \multicolumn{3}{c|}{Private 1}                & \multicolumn{3}{c|}{Private 2}                & \multicolumn{3}{c|}{ASAP 1}                   & \multicolumn{3}{c|}{ASAP 2}                   & \multicolumn{3}{c|}{ASAP 5}                   & \multicolumn{3}{c|}{ASAP 6}                   & \multicolumn{3}{c}{Overall} \\ \hline
\multicolumn{2}{c|}{Eval Metrics}                                   & Acc    & F1     & \multicolumn{1}{c|}{QWK}    & Acc    & F1     & \multicolumn{1}{c|}{QWK}    & Acc    & F1     & \multicolumn{1}{c|}{QWK}    & Acc    & F1     & \multicolumn{1}{c|}{QWK}    & Acc    & F1     & \multicolumn{1}{c|}{QWK}    & Acc    & F1     & \multicolumn{1}{c|}{QWK}    & Acc     & F1      & QWK     \\ \hline
Model                         & \multicolumn{1}{c|}{Stage}          & \multicolumn{21}{c}{Trained on All Data}                                                                                                                                                                                                                                                                                    \\ \hline
\multirow{2}{*}{LLaMA 3 8B}   & \multicolumn{1}{c|}{SFT}            & 0.5394 & 0.5180 & \multicolumn{1}{c|}{0.8118} & 0.5306 & 0.5313 & \multicolumn{1}{c|}{0.7548} & 0.7058 & 0.7142 & \multicolumn{1}{c|}{0.8360} & 0.6643 & 0.6769 & \multicolumn{1}{c|}{0.7925} & 0.8595 & 0.5921 & \multicolumn{1}{c|}{\underline{0.7936}} & 0.8381 & 0.6079 & \multicolumn{1}{c|}{0.7628} & 0.6896  & 0.6067  & 0.7919  \\
                              & \multicolumn{1}{c|}{DPO}            & 0.5315 & 0.5226 & \multicolumn{1}{c|}{\underline{0.8439}} & 0.5459 & 0.5570 & \multicolumn{1}{c|}{\underline{0.7624}} & 0.7527 & 0.7616 & \multicolumn{1}{c|}{\underline{0.8579}} & 0.6408 & 0.6525 & \multicolumn{1}{c|}{\underline{\textbf{0.8038}}} & 0.8612 & 0.6364 & \multicolumn{1}{c|}{0.7735} & 0.8715 & 0.5809 & \multicolumn{1}{c|}{\underline{0.8042}} & 0.7006  & 0.6185  & \underline{0.8076}  \\ \hline
\multirow{2}{*}{Mixtral 8$\times$7B} & \multicolumn{1}{c|}{SFT}            & 0.5906 & 0.5799 & \multicolumn{1}{c|}{0.8326} & 0.5255 & 0.5213 & \multicolumn{1}{c|}{0.7324} & 0.7310 & 0.7398 & \multicolumn{1}{c|}{0.8511} & 0.5915 & 0.5837 & \multicolumn{1}{c|}{0.6886} & 0.8579 & 0.5948 & \multicolumn{1}{c|}{0.8174} & 0.8564 & 0.5942 & \multicolumn{1}{c|}{0.7735} & 0.6922  & 0.6023  & 0.7826  \\
                              & \multicolumn{1}{c|}{DPO}            & 0.6063 & 0.6016 & \multicolumn{1}{c|}{\underline{\textbf{0.8440}}} & 0.5867 & 0.5917 & \multicolumn{1}{c|}{\underline{\textbf{0.7802}}} & 0.7744 & 0.7808 & \multicolumn{1}{c|}{\underline{\textbf{0.8749}}} & 0.6526 & 0.6763 & \multicolumn{1}{c|}{\underline{0.7788}} & 0.8779 & 0.6326 & \multicolumn{1}{c|}{\underline{\textbf{0.8447}}} & 0.8715 & 0.6011 & \multicolumn{1}{c|}{\underline{\textbf{0.8245}}} & 0.7282  & 0.6473  & \underline{\textbf{0.8245}}  \\ \hline
\multicolumn{23}{c}{Trained on All Data w/o ASAP2}                                                                                                                                                                                                                                                                                                                                                \\ \hline
\multirow{2}{*}{LLaMA 3 8B}   & \multicolumn{1}{c|}{SFT} & 0.5197 & 0.5079 & \multicolumn{1}{c|}{0.7876} & 0.5255 & 0.5203 & \multicolumn{1}{c|}{0.7581} & 0.6733 & 0.6834 & \multicolumn{1}{c|}{0.8096} & \cellcolor{blue!25}0.3615 & \cellcolor{blue!25}0.3397 & \multicolumn{1}{c|}{\cellcolor{blue!25}0.4396} & 0.8512 & 0.6235 & \multicolumn{1}{c|}{0.7389} & 0.8214 & 0.5647 & \multicolumn{1}{c|}{0.7394} & 0.6254  & 0.5399  & 0.7122  \\
                              & \multicolumn{1}{c|}{DPO} & 0.4803 & 0.4459 & \multicolumn{1}{c|}{\underline{0.8256}} & 0.5561 & 0.5685 & \multicolumn{1}{c|}{\underline{\textbf{0.7658}}} & 0.7310 & 0.7417 & \multicolumn{1}{c|}{\underline{0.8536}} & \cellcolor{blue!25}0.4131 & \cellcolor{blue!25}0.3854 & \multicolumn{1}{c|}{\underline{\cellcolor{blue!25}0.5117}} & 0.8495 & 0.5562 & \multicolumn{1}{c|}{\underline{0.7413}} & 0.8798 & 0.5881 & \multicolumn{1}{c|}{\underline{0.8087}} & 0.6517  & 0.5476  & \underline{0.7511}  \\ \hline
\multirow{2}{*}{Mixtral 8$\times$7B} & \multicolumn{1}{c|}{SFT}            & 0.5866 & 0.5756 & \multicolumn{1}{c|}{0.8311} & 0.5102 & 0.4991 & \multicolumn{1}{c|}{0.7156} & 0.7040 & 0.7143 & \multicolumn{1}{c|}{0.8386} & \cellcolor{blue!25}0.5164 & \cellcolor{blue!25}0.5217 & \multicolumn{1}{c|}{\cellcolor{blue!25}\underline{\textbf{0.6744}}} & 0.8545 & 0.6036 & \multicolumn{1}{c|}{0.8055} & 0.8414 & 0.5623 & \multicolumn{1}{c|}{0.7648} & 0.6689  & 0.5794  & 0.7717  \\
                              & \multicolumn{1}{c|}{DPO}            & 0.6378 & 0.6388 & \multicolumn{1}{c|}{\underline{\textbf{0.8535}}} & 0.5357 & 0.5446 & \multicolumn{1}{c|}{\underline{0.7587}} & 0.7726 & 0.7802 & \multicolumn{1}{c|}{\underline{\textbf{0.8727}}} & \cellcolor{blue!25}0.5141 & \cellcolor{blue!25}0.5250 & \multicolumn{1}{c|}{\cellcolor{blue!25}0.6589} & 0.8696 & 0.6054 & \multicolumn{1}{c|}{\underline{\textbf{0.8295}}} & 0.8648 & 0.5703 & \multicolumn{1}{c|}{\underline{\textbf{0.8129}}} & 0.6991  & 0.6107  & \underline{\textbf{0.7977}}  \\ \hline
\multicolumn{23}{c}{Trained on All Data w/o Private data}                                                                                                                                                                                                                                                                                                                                         \\ \hline
\multirow{2}{*}{LLaMA 3 8B}   & \multicolumn{1}{c|}{SFT}            & \cellcolor{blue!25}0.3701 & \cellcolor{blue!25}0.3617 & \multicolumn{1}{c|}{\cellcolor{blue!25}0.6839} & \cellcolor{blue!25}0.3367 & \cellcolor{blue!25}0.2894 & \multicolumn{1}{c|}{\underline{\cellcolor{blue!25}0.3982}} & 0.6426 & 0.6589 & \multicolumn{1}{c|}{0.8058} & 0.6338 & 0.6495 & \multicolumn{1}{c|}{0.7577} & 0.8495 & 0.6090 & \multicolumn{1}{c|}{\underline{0.7764}} & 0.8331 & 0.5776 & \multicolumn{1}{c|}{0.7434} & 0.6110 & 0.5243 & 0.6942  \\
                              & \multicolumn{1}{c|}{DPO}            & \cellcolor{blue!25}0.4331 & \cellcolor{blue!25}0.3944 & \multicolumn{1}{c|}{\cellcolor{blue!25}\underline{\textbf{0.7457}}} & \cellcolor{blue!25}0.2908 & \cellcolor{blue!25}0.2096 & \multicolumn{1}{c|}{\cellcolor{blue!25}0.3503} & 0.7112 & 0.7210 & \multicolumn{1}{c|}{\underline{0.8343}} & 0.6526 & 0.6659 & \multicolumn{1}{c|}{\underline{\textbf{0.7953}}} & 0.8545 & 0.5617 & \multicolumn{1}{c|}{0.7601} & 0.8648 & 0.5950 & \multicolumn{1}{c|}{\underline{0.8071}} & 0.6345 & 0.5246 & \underline{0.7155} \\ \hline
\multirow{2}{*}{Mixtral 8$\times$7B} & \multicolumn{1}{c|}{SFT}            & \cellcolor{blue!25}0.3937 & \cellcolor{blue!25}0.3832 & \multicolumn{1}{c|}{\cellcolor{blue!25}0.6594} & \cellcolor{blue!25}0.5051 & \cellcolor{blue!25}0.5020 & \multicolumn{1}{c|}{\cellcolor{blue!25}\underline{\textbf{0.6819}}} & 0.7130 & 0.7237 & \multicolumn{1}{c|}{0.8439} & 0.5822 & 0.5755 & \multicolumn{1}{c|}{\underline{0.6920}} & 0.8428 & 0.5460 & \multicolumn{1}{c|}{0.7942} & 0.8497 & 0.5834 & \multicolumn{1}{c|}{0.7841} & 0.6478 & 0.5523 & \underline{\textbf{0.7426}}  \\
                              & \multicolumn{1}{c|}{DPO}            & \cellcolor{blue!25}0.4606 & \cellcolor{blue!25}0.4603 & \multicolumn{1}{c|}{\cellcolor{blue!25}\underline{0.7060}} & \cellcolor{blue!25}0.4592 & \cellcolor{blue!25}0.4395 & \multicolumn{1}{c|}{\cellcolor{blue!25}0.6120} & 0.7563 & 0.7654 & \multicolumn{1}{c|}{\underline{\textbf{0.8643}}} & 0.5258 & 0.5562 & \multicolumn{1}{c|}{0.6329} & 0.8679 & 0.6151 & \multicolumn{1}{c|}{\underline{\textbf{0.8272}}} & 0.8748 & 0.6220 & \multicolumn{1}{c|}{\underline{\textbf{0.8090}}} & 0.6574  & 0.5764  & 0.7419 \\\hline
\multicolumn{23}{c}{Ablation Study on Decoding Method (Trained on All Data)}                                                                                                                                                                                                                                                                                                                                         \\ \hline
LLaMA 3 8B   & \multicolumn{1}{c|}{SFT}            & 0.5433 & 0.5247 & \multicolumn{1}{c|}{0.7944} & 0.5357 & 0.5342 & \multicolumn{1}{c|}{0.7421} & 0.6733 & 0.6870 & \multicolumn{1}{c|}{0.8247} & 0.6526 & 0.6647 & \multicolumn{1}{c|}{0.7757} & 0.8462 & 0.6043 & \multicolumn{1}{c|}{0.7741} & 0.8397 & 0.6106 & \multicolumn{1}{c|}{0.7434} & 0.6522 & 0.5505 & 0.7543  \\
Mixtral 8$\times$7B & \multicolumn{1}{c|}{SFT}            & 0.5827 & 0.5590 & \multicolumn{1}{c|}{0.8206} & 0.5000 & 0.4916 & \multicolumn{1}{c|}{0.7151} & 0.6426 & 0.6559 & \multicolumn{1}{c|}{0.7999} & 0.4953 & 0.4797 & \multicolumn{1}{c|}{0.6303} & 0.8411 & 0.5157 & \multicolumn{1}{c|}{0.7929} & 0.8514 & 0.6011 & \multicolumn{1}{c|}{0.7667} & 0.7076 & 0.5631 & 0.7475 \\ \hline
\end{tabular}}
    \caption{Overall Comparison of Assessment Performance under Three Different Combinations of Training Data Settings. The highest QWK under each training dataset setting is highlighted in \textbf{Bold}, and the highest QWK under the same base model has been \underline{Underlined}. \colorbox{blue!25}{Colored cells} indicates the testing results on unseen datasets.}
    \label{tab:further_assessment_results}
\end{sidewaystable*}

\begin{figure}[h]
    \centering
    \begin{minipage}{0.25\linewidth}
        \centering
        \includegraphics[width=1.0\linewidth]{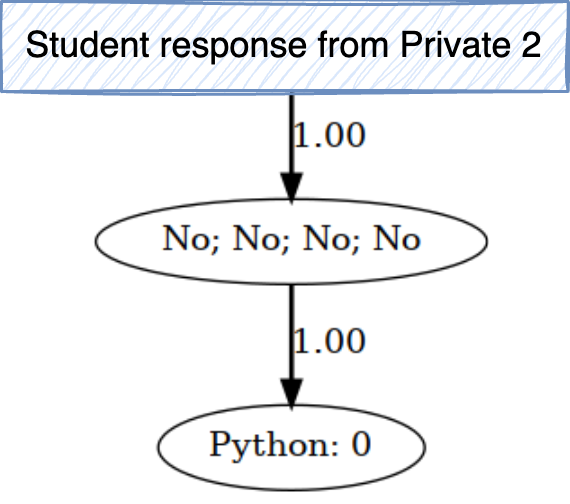} %
    \end{minipage}\hfill
    \begin{minipage}{0.75\linewidth}
        \centering
        \includegraphics[width=1.1\linewidth]{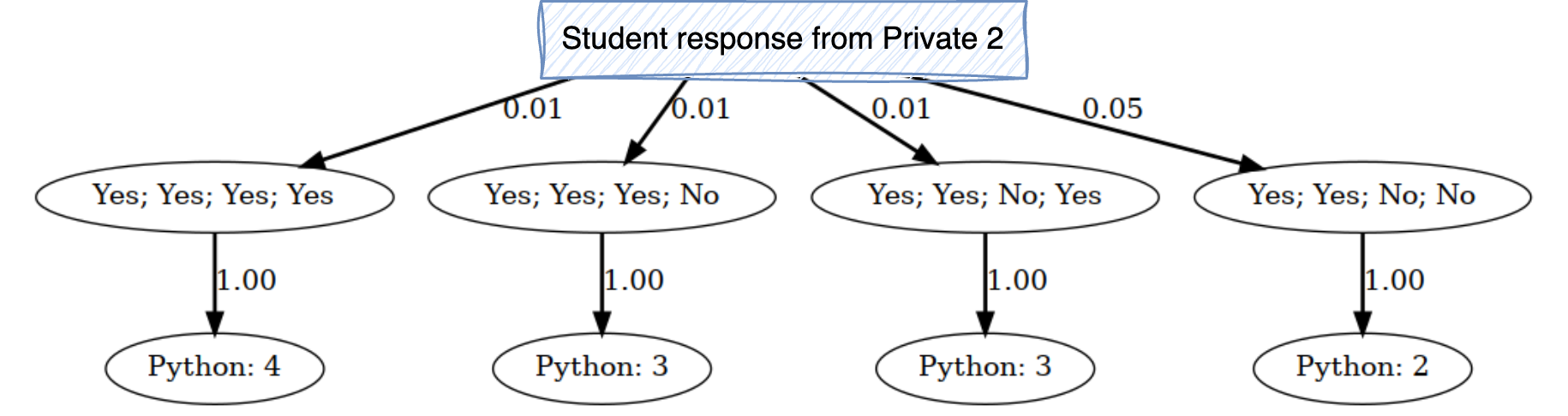} %
    \end{minipage}
    \caption{Comparison between Thought Tree Before (left) and After (right) Training with Our Framework.}
    \label{fig:tree_compare}
\end{figure}

\subsection{Case Study: Correction on Thought Tree}
As illustrated in Figure \ref{fig:tree_compare}, although our model was not trained on thought trees, we observe that the highest probability path ($5\%$ shown on the right-hand side) accurately assesses the score. The path ``\textit{Yes; Yes; No; No}'' indicates that the student's response covered the first two key answer elements but missed the last two. The rationale generated for this student's response also reflects this decision by validating the first two key answer elements. However, due to finetuning, we can no longer generate a thought tree for some responses. Nonetheless, for those that can still be generated, we have identified patterns that suggest our trained model could potentially calibrate the LLM to generate thought trees that are coherent with its rationale.

\subsection{LLM Automated Evaluation}
\subsubsection{Prompt Template}
Following the DPO \cite{dpo}, we designed the following prompt template to evaluate the rationale generated from our framework and calculate the win rate.
\begin{tcolorbox}[title=Prompt Template for Response Level Rationale Summarization for GPT-4,
    colback=white,
    colframe=yellow!75!black,
    colbacktitle=yellow,
    coltitle=black,
    breakable,
    label=tcb:win_rate_prompt,
    fonttitle=\bfseries]
For the following assessment rationale for the student response, which one is more accurate and helpful?\\
\text{[Question]}: \{question\}\\
\text{[Key Answer Elements]}: \{key\_answer\_elements\}\\
\text{[Marking Rubric]}: \{marking\_rubric\}\\
\text{[Student Response]}: "\{student\_response\}"\\
\text{[Teacher Assessed Mark]}: \{mark\}\\
\\
Rationale A:\\
"\{baseline\}"\\
Rationale B:\\
"\{response\}"\\
FIRST provide a one-sentence comparison of the two rationales and explain which you feel is more accurate and helpful. SECOND, on a new line, state only "A" or "B" to indicate which response is more accurate and helpful. Your response should use the format:\\
Comparison: <one-sentence comparison and explanation>\\
More accurate and helpful: <"A" or "B">
\end{tcolorbox}
\subsubsection{Sampled Win Rate Rationales}
As shown in Table \ref{tab:win_rate_sampled}, we sampled some cases of GPT-4's win rate assessments, complete with rationales and decisions.
\begin{table*}[!h]
\centering
\resizebox{\linewidth}{!}{
\begin{tabular}{p{\linewidth}}
\hline
\small \textbf{Student Response} (from ASAP 6): Ground-truth mark: 0\\
\small
mRNA will change into tRNA and will leave amino acids. Then the amino acids will make a ribosome. mRNA will send a message to tRNA. Finally ribosomes will connect and make a bond\\
\small \textbf{A: AERA Generated Rationale}: \\
\small 1 point; The student answer matches one key element, '...mRNA is sent to the ribosomes...'. However, the other statements are incorrect or incomplete.\\
\small \textbf{B: Mixtral 8$\times$7B SFT Generated Rationale}: \\
\small The student's answer scored 0 points according to the marking rubric which awards 3 points for addressing four key elements, 2 points for three key elements, 1 point for one or two key elements, and 0 points for other responses. The student's answer contains several fundamental errors and misunderstandings about the protein synthesis process. Key inaccuracies include stating'mRNA will change into tRNA' and 'the amino acids will make a ribosome,' which demonstrate a confusion between the roles and identities of mRNA, tRNA, and ribosomes. The student also fails to describe critical steps such as mRNA exiting the nucleus, mRNA traveling to the ribosome, the reading of mRNA codons by rRNA, and the specific interactions and roles of tRNA in carrying and adding amino acids. None of the key answer elements were correctly addressed as required by the marking rubric.\\
\small \textbf{GPT-4 Evaluation}:\\
\small Comparison: Rationale B is more detailed and accurate, providing specific feedback on the student's misunderstandings and errors, which is essential for learning, whereas Rationale A is overly generous and lacks detail on the specific errors made.\\
\small More accurate and helpful: B\\
\hline
\small
\textbf{Student Response} (from ASAP 2): Ground-truth mark: 0\\
\small
The pour into the take the vinegar in each containers.  who is the problem ( Hypothesis)  How you can do this experiment  A table who explain the experiment  use more types of masses like glass, paper, etc.\\
\small \textbf{A: AERA Generated Rationale}: \\
\small
1 point; This response describes one additional piece of information that would be needed to accurately replicate the experiment: 'the pour into the water in each container.'\\
\small \textbf{B: Mixtral 8$\times$7B DPO Generated Rationale}: \\
\small TThe student's answer scored 0 points as per the marking rubric, which requires the response to describe additional pieces of information needed to accurately replicate the experiment. The student's answer failed to specify any necessary details such as the amount of vinegar used, the type of vinegar, the materials tested, the size/surface area of materials, the rinsing duration, the drying time, the drying method, or the size/type of container. The student's response included vague and unrelated statements such as 'The pour into the take the vinegar in each containers. who is the problem ( Hypothesis) How you can do this experiment A table who explain the experiment use more types of masses like glass, paper, etc.' which do not address the specifics required for replication.\\
\small \textbf{GPT-4 Evaluation}:\\
\small Comparison: Rationale B is more accurate and helpful because it clearly explains why the student's response failed to meet the requirements of the marking rubric by not providing specific and relevant information needed to replicate the experiment, whereas Rationale A incorrectly credits the student for providing useful information.\\
\small More accurate and helpful: B\\
\hline
\end{tabular}}
\caption{Sampled GPT-4 Win Rate with Evaluated Preference Decisions and Rationales.}
\label{tab:win_rate_sampled}
\end{table*}

\subsection{Human Evaluation} \label{sec:human_eval}

\subsubsection{Evaluation Setup}
We conducted our human evaluation using the Prolific\footnote{\url{https://app.prolific.com/}} annotation platform. To ensure efficient evaluation, we divided the sampled data into batches, each containing sample outputs from the same question. This allowed for a consistent evaluation standard across all rationale assessments. All annotators were screened to ensure they had higher education from an English-speaking country. We compensated our annotators at a rate of \$12.80 per hour.

\subsubsection{Evaluation on Rationale Quality}

\begin{figure*}[t]
  \centering
  \includegraphics[width=\linewidth]{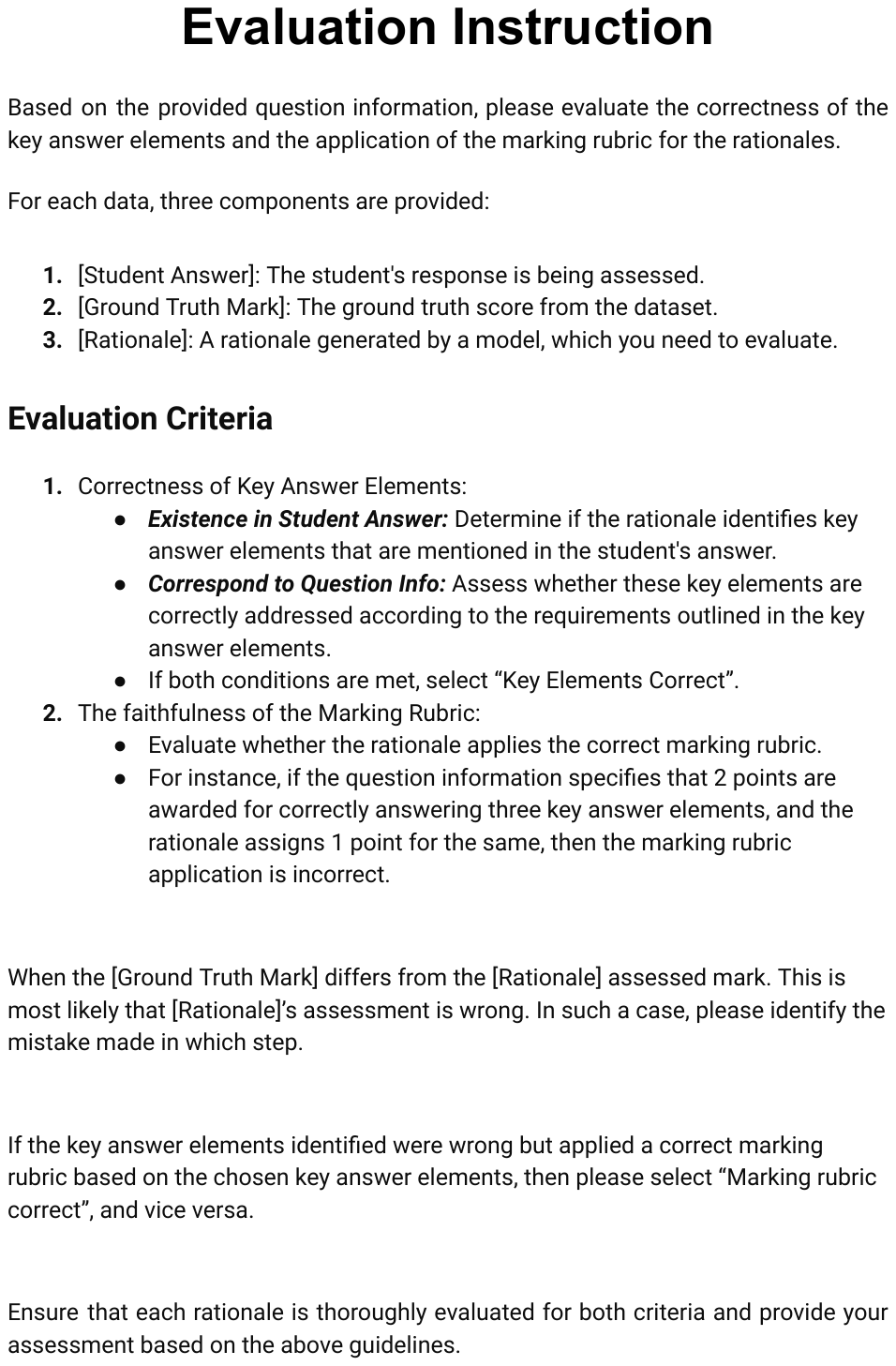}
  \caption{Evaluation Instruction Provided to the Anonymous Annotators.}
  \label{fig:annotation_instruction}
\end{figure*}

\paragraph{Rationale Evaluation Instruction}
The detailed annotation instruction given to our annotator is presented in Figure \ref{fig:annotation_instruction}.

\subsubsection{Evaluation on Stage 2 Summarization Faithfulness}

\begin{table}[!h]
\centering
\resizebox{\linewidth}{!}{%
\small{
\begin{tabular}{cc}
\toprule
\textbf{Evaluated Aspects} & \textbf{Accuracy} \\ 
\midrule
Scoring Coherent with Tree Path & 93.33\%\\
Coherence Between Mark and Rationale & 100\% \\
Correctness on Summarize Key Elements & 96.67\% \\
Correctness on Apply Marking Rubric & 100\% \\
\bottomrule
\end{tabular}}}
\caption{Human Evaluation on Summarization Faithfulness.}
\label{tab:human_evaluation_summarization}
\end{table}

In Stage 2 of our framework, we model the response-level synthetic rationale generation as a summarization task. In this section, we present a human evaluation of the faithfulness of the generated rationales. We evaluated the faithfulness across four aspects with binary annotation from a total of 30 sampled cases.

The first aspect is ``Scoring alignment with tree path,'' which assesses whether the assigned mark matches the tree path's score. We found that in 93.33\% of the sampled cases, the mark column aligns with the tree path, indicating reliable summarization. Similarly, since we used the mark and rationale columns to train a score extractor, we evaluated whether the mark corresponds to the assessment decision presented in the rationale. We found that all marks are correctly aligned with the rationale's result.

To evaluate the faithfulness of the content, we focused on the key answer elements and the application of marking rubrics. The evaluation does not consider the correctness of the assessment but only whether the output is faithful to the given input. We found no cases among the samples where an incorrect marking rubric was used to conclude a score. Only one case showed misalignment with the key answer element. In this instance, the original input decided ``Yes'', indicating that the student answered the corresponding key answer element, but the output rationale stated that the quoted part was too vague to be counted as a correct answer.

\subsection{Error Analysis}
As shown in Table \ref{tab:erroneous}, we present some error analysis on the generated rationales. We found that the model sometimes struggled to explicitly define the scope of key answer elements. 

For the first case, the student answered ``amount of distilled water being used to rinse each sample,'' which can be identified as a possible correct answer to the rinsing step. However, the model did not acknowledge this as a relevant element for rinsing duration, leading to an incomplete assessment of the student's response.

Similarly, for the second case, the student's answer on ``allowing the weights to hang longer to see if the plastic would stretch more'' indirectly relates to ensuring consistent experimental conditions. The rationale did not fully recognize this improvement suggestion, missing the connection to maintaining consistency in experimental conditions over time.

In the third case, the student's descriptions of osmosis, endocytosis, and exocytosis were correct and matched the key elements regarding these processes. Yet, the rationale overlooked the accuracy of these descriptions and emphasized the absence of additional details, resulting in an underappreciation of the student's correct answers.

These examples indicate that the model's rationale generation occasionally fails to fully recognize and credit partially correct elements in student answers, leading to potentially incorrect assessment.

\begin{table*}[!h]
\centering
\resizebox{\linewidth}{!}{
\begin{tabular}{p{\linewidth}}
\hline
\small [Student Answer]: After reading the groups procedure, I would need to know the names of the three samples being tested in the experiment. The amount of the sample being poured into the containers would also be necessary as well as the amount of distilled water being used to rinse each sample. Another important factor would be to know the location of samples were placed in to sit and dry for 30 minutes. How the students recorded the mass would also be needed to replicate the experiment.\\
\small  [Ground Truth Mark]: 2 \\
\small  [Rationale]: The student's answer scored 1 point according to the marking rubric, as it describes one additional piece of information needed to accurately replicate the experiment. The student correctly identified the need for specifying '...the amount of the sample being poured into the containers...' which aligns with knowing the quantity of vinegar used. However, other key answer elements such as the type of vinegar, the materials to test, size\/surface area of materials, \textcolor{red}{rinsing duration}, drying time, drying method, and container size\/type were not addressed. The student's response did not mention these elements, which are vital for replicating the experiment accurately. \\
\hline
\small
\small [Student Answer]: Based on the students data I can conclude that plastic type B has the largest amount stretched. Two ways that the student could have improved the experimental decison is by allowing the weights to hang longer to see if the plastic would stretch more. Another way he could have improved his experiment is by doing a third trials to really be sure that his results are correct. \\
\small [Ground Truth Mark]: 2 \\
\small [Rationale]: The student's answer scored 3 points as per the marking rubric, which is applied for drawing a valid conclusion supported by data and describing two ways to improve the experimental design or validity. Specifically, the student concluded that 'plastic type B has the largest amount stretched,' matching the key answer element of drawing a valid experimental conclusion. Additionally, the student proposed performing additional trials ('...doing a third trials to really be sure that his results are correct') and allowing more time for weights to hang, \textcolor{red}{which indirectly relates to ensuring consistent experimental conditions}, both suggesting improvements to the experimental design. However, the answer lacked suggestions on standardizing sample sizes or thicknesses and did not clarify control over variables such as the weight amounts, which could further enhance the experiment's validity. \\
\hline
\small [Student Answer]: Osmosis, Is the movement of water across the membrain. Endocitosis, is the movement of things in to the cell. Exocitosis, is the movement of things out of the cell.\\
\small [Ground Truth Mark]: 1 \\
\small [Rationale]: The student's answer scored 2 points according to the marking rubric, which awards 3 points for three key elements, 2 points for two key elements, 1 point for one key element, and 0 points for none. The student correctly described osmosis as 'the movement of water across the membrane' and endocytosis as 'the movement of things in to the cell' and exocytosis as 'the movement of things out of the cell', \textcolor{red}{matching three key elements regarding descriptions of osmosis, endocytosis, and exocytosis}. However, the answer lacked descriptions of selective permeability, passive transport, facilitated diffusion, active transport, use of pumps, and various specifics of membrane-assisted transport (like phagocytosis and pinocytosis), and did not mention protein channels.\\
\hline
\end{tabular}}
\caption{Sampled Erroneous Rationales Generated from Mixtral 8$\times$7B DPO.}
\label{tab:erroneous}
\end{table*}

\subsection{Further Case Studies}

\begin{table*}[!h]
\centering
\resizebox{\linewidth}{!}{%
\begin{tabular}{p{1.1\linewidth}}
\hline
\small
\textbf{Student Response} (from ASAP 1): In order to replicate the experiment there needs to be \textcolor{red}{how much vinegar to use}. \\
\small \textbf{Ground-truth mark}: 1\\
\small \textbf{AERA}: 1 point; This response describes one additional piece of information that would be needed to accurately replicate the experiment: \textcolor{red}{“how much vinegar to use.”}\\
\small \textbf{Mixtral 8$\times$7B SFT}: The student's answer received a score of 1 based on the marking rubric, which awards \textcolor{red}{1 point for describing one additional piece of information needed to accurately replicate the experiment}. The student's response, \textcolor{red}{'In order to replicate the experiment there needs to be how much vinegar to use,' correctly identifies the need to specify the amount of vinegar used,} which is essential for replicating the experiment. However, the response fails to address other key elements such as the type of vinegar, the materials to test, the size\/surface area of materials, the rinsing duration, the drying time, the drying method, and the size\/type of container. These omissions led to the awarding of only 1 point.\\
\small \textbf{Mixtral 8$\times$7B DPO}: The student's answer received a score of 1 according to the marking rubric, which awards \textcolor{red}{1 point for describing one additional piece of information needed to accurately replicate the experiment}. The student's answer mentioned the need to specify \textcolor{red}{'how much vinegar to use,' which aligns with one of the Key Answer Elements}. However, the answer did not address other essential information such as the type of vinegar, the materials to test, the size\/surface area of materials, the rinsing duration, drying time, drying method, or the size/type of container. Therefore, only 1 point was awarded as the response only partially met the criteria.\\
\hline
\small\textbf{Student Response} (from ASAP 2): A conclusion that can be dawn from the students experiment is that plastic B is able to stretch the farthest (T1:22mm, T2:23mm) and plastic A stretched the least (T1:10mm, T2:12mm). The validity of this experiment could have been improved greatly by doing these two things; 1)Making sure that the starting length of the plastics were all the same and 2)determining how much of the plastic will be stretched to the table at the beginning. By doing these two things, you're ensuring controlled variables therefore validating your experiment\\
\small \textbf{Ground-truth mark}: 3\\
\small \textbf{AERA}: 3 points; The student provides \textcolor{red}{an acceptable conclusion based on the data: “...plastic B is able to stretch the farthest....” Two correct ways to improve the experimental design and/or the validity of the results are given: “...making sure that the starting length of the plastics were all the same...” and “...specify how much of the plastic will be stretched to the table at the beginning.”}\\
\small \textbf{Mixtral 8$\times$7B SFT}: The student's answer scored 3 points according to the marking rubric, as it both \textcolor{red}{draws a valid conclusion supported by the data and describes two ways to improve the experimental design and\/or validity of the results}. The student \textcolor{red}{correctly concludes that 'plastic B is able to stretch the farthest... and plastic A stretched the least'}, aligning with the key answer element for drawing a valid experimental conclusion. Additionally, the student \textcolor{red}{suggests ensuring 'the starting length of the plastics were all the same' and determining 'how much of the plastic will be stretched to the table at the beginning'}, which addresses the key elements of controlling variables and ensuring uniformity in sample sizes. These suggestions directly improve the experimental design and validity, fulfilling the requirements for a full score.\\
\small \textbf{Mixtral 8$\times$7B DPO}: The student's answer scored 3 points as per the marking rubric, which requires \textcolor{red}{drawing a valid conclusion supported by data and describing two ways to improve the experimental design or validity}. The student correctly \textcolor{red}{concluded that 'plastic B is able to stretch the farthest... and plastic A stretched the least'}, matching the key element of a valid experimental conclusion. \textcolor{red}{Improvements suggested included ensuring 'the starting length of the plastics were all the same', addressing the need for consistent initial measurements}. However, the student did not suggest ensuring the same thickness or the same number of weights, which would control more variables but were not crucial to achieve the full mark. The student's answer adequately addressed two experimental design improvements.\\
\hline 
\small\textbf{Student Response} (from ASAP 5): The first step is when the mRNA attaches to the ribosomes, which are produced by rRNA. The mRNA forms a codon. Then, tRNA anticodons bring amino acids to match the codons. The amino acids attach by peptide bonds. After releasing the amino acid, the tRNA leaves to get more amino acids. The process stops when a complete protein is made.\\
\small \textbf{Ground-truth mark}: 2\\
\small \textbf{AERA}: 3 points; The student answer matches four key elements, “...mRNA attaches to a ribosomes...”, “...tRNA attaches to the ribosomes...”, “...tRNA brings amino acids that matches with the ribosomes...”, “...tRNA bring amino acids and pairs with codons...” and “...amino acids form a polypeptide chain...”.\\
\small \textbf{Mixtral 8$\times$7B SFT}: The student's answer scored 3 points according to the marking rubric, which awards \textcolor{red}{3 points for addressing four key elements}. The student successfully \textcolor{red}{described the mRNA traveling to the ribosome ('The mRNA attaches to the ribosomes'), the reading of mRNA bases in triplets called codons ('The mRNA forms a codon'), tRNA carrying the complementary anticodon recognizing the complementary codon of the mRNA ('tRNA anticodons bring amino acids to match the codons'), and the bonding of amino acids on tRNA ('The amino acids attach by peptide bonds')}. However, the answer lacked explicit mention of mRNA exiting the nucleus via nuclear pore, the addition of new amino acids to tRNA, the initiation process at the START codon, and the elongation and termination at the STOP codon.\\
\small \textbf{Mixtral 8$\times$7B DPO}: The student's answer was awarded \textcolor{red}{2 points} as per the marking rubric that assigns this score for identifying \textcolor{red}{three key elements in their response}. Specifically, the student \textcolor{red}{correctly described'mRNA leaving the nucleus' implicity when stating the mRNA 'attaches to the ribosomes', described tRNA carrying the complementary anticodon recognizing the mRNA codon as indicated by 'tRNA anticodons bring amino acids to match the codons', and addressed the bonding of amino acids through 'The amino acids attach by peptide bonds.'} However, the response lacked explicit mention of mRNA traveling through the cytoplasm, reading of mRNA bases in triplets by rRNA, the process of a new amino acid being added to the tRNA, and the specifics of initiation and termination including START and STOP codons, hence losing potential points.\\
\hline
\small\textbf{Student Response} (from ASAP 6): Osmosis controls the water movement across the cell membrane. The cell membrane allows for substances that the cell needs in, and gets rid of the waste and unwanted material.\\
\small \textbf{Ground-truth mark}: 1\\
\small \textbf{AERA}: 1 point; This student answer matches one key element, ""Osmosis... movement of water across the cell membrane"". The other two explanations are incorrect and incomplete.\\
\small \textbf{Mixtral 8$\times$7B SFT}: The student's answer scored 1 point according to the marking rubric, which \textcolor{red}{awards 1 point for addressing one key answer element}. The student successfully \textcolor{red}{described osmosis as a process that 'controls the water movement across the cell membrane,' aligning with the key answer element that osmosis is the diffusion of water across the cell membrane}. However, the answer failed to address other key elements such as selective permeability, passive transport, facilitated diffusion, active transport, and various forms of membrane-assisted transport (exocytosis, endocytosis, phagocytosis, pinocytosis), and the use of protein channels or pumps.\\
\small \textbf{Mixtral 8$\times$7B DPO}: The student's answer received a score of 1 according to the marking rubric, which awards \textcolor{red}{1 point for addressing one key element correctly.} The student \textcolor{red}{successfully described osmosis as the process that 'controls the water movement across the cell membrane,'} aligning with the key element on osmosis as the diffusion of water. However, the answer failed to address other key elements such as selective permeability, passive transport, facilitated diffusion, active transport, and various aspects of membrane-assisted transport (exocytosis, endocytosis, phagocytosis, pinocytosis), and the use of protein channels or pumps.\\
\hline
\end{tabular}}
\caption{More case studies. Critical statements from the rationales are highlighted in \textcolor{red}{red}.}
\label{tab:app_case_study}
\end{table*}

In this section, we provide further case studies on the generated rationale from AERA, Mixtral 8$\times$7B SFT, and Mixtral 8$\times$7B DPO models, as shown in Table \ref{tab:app_case_study}. 

In the first case from ASAP 1, the ground-truth mark is 1. The AERA model correctly identifies that the student specified the amount of vinegar needed, matching one of the required key elements. Similarly, both the SFT and DPO models accurately highlight the student's recognition of the need to quantify the vinegar, aligning with the marking rubric. However, the SFT and DPO models elaborate on the omission of other experimental details, providing more comprehensive feedback.

The second case from ASAP 2 has a ground-truth mark of 3. The AERA model correctly credits the student for drawing a valid conclusion based on the data and suggesting two improvements to the experimental design. The SFT and DPO models also recognize the same elements, with SFT providing detailed feedback on the student's correct conclusions and suggestions. The DPO model mirrors the SFT model's rationale but emphasizes the importance of consistent initial measurements, showcasing the DPO's ability to reinforce key points concisely.

In the third case from ASAP 5, the ground-truth mark is 2. The AERA model overestimates the student's response and provides very vague rationales, awarding 3 points for matching four key elements. In contrast, the SFT model identifies three key elements and provides detailed explanations, addressing the student's accurate descriptions of mRNA, tRNA, and peptide bonds. The DPO model could improve its assessment and give a score of 2 points, recognizing only three key elements and noting the lack of explicit mention of certain steps in the process.

The fourth case from ASAP 6 has a ground-truth mark of 1. The AERA model accurately matches the student's response to one key element about osmosis. Both the SFT and DPO models similarly award 1 point, providing detailed feedback on the accurate description of osmosis while noting the omission of other critical elements related to cell membrane transport processes.

Overall, these case studies highlight the ability of the Mixtral 8$\times$7B SFT and DPO models to provide detailed and accurate rationales, often improving upon the AERA model's outputs by offering more specific feedback and better alignment with the marking rubric. The DPO model, in particular, excels in conciseness while maintaining the necessary detail for accurate assessment.

\end{document}